\documentclass[journal]{IEEEtran}
\usepackage{graphicx}
 \usepackage{float}
\usepackage[caption=false]{subfig}

\usepackage{amsmath,amsfonts}
\usepackage{algorithm, algorithmic}
\usepackage{array}

\usepackage{textcomp}
\usepackage{stfloats}
\usepackage{url}
\usepackage{verbatim}

\usepackage{mathrsfs}
\usepackage{bm}
\usepackage{amssymb}
\usepackage{amsmath}

\usepackage{caption}
\usepackage{hyperref}
\usepackage{booktabs}
\usepackage{color,colortbl}
\usepackage{dsfont}
\usepackage{multirow}

\definecolor{orange}{RGB}{200, 75, 49}

\def\BibTeX{{\rm B\kern-.05em{\sc i\kern-.025em b}\kern-.08em
    T\kern-.1667em\lower.7ex\hbox{E}\kern-.125emX}}
\usepackage{balance}
\begin{document}
\title{HHF: Hashing-guided Hinge Function for Deep Hashing Retrieval}
\author{   
    Chengyin Xu$^*$,
    Zenghao Chai$^*$,
    Zhengzhuo Xu$^*$,\\
    Hongjia Li,
    Qiruyi Zuo,
    Lingyu Yang,
    Chun Yuan$^\dagger$, \textit{Senior Member, IEEE}
    \thanks{$^\dagger$Corresponding author: C. Yuan is with the Tsinghua Shenzhen International Graduate School, Tsinghua University, Shenzhen 518055, China. (email: yuanc@sz.tsinghua.edu.cn).
    
    $^*$Equal contribution authors, listing order is random: C. Xu, Z. Chai and Z. Xu are with the Tsinghua Shenzhen International Graduate School, Tsinghua University, Shenzhen 518055, China. (e-mail: \{xucy20, xzz20\}@mails.tsinghua.edu.cn, zenghaochai@gmail.com).
    
    H. Li, Q. Zuo, and L. Yang are with the Tsinghua Shenzhen International Graduate School, Tsinghua University, Shenzhen 518055, China. (e-mail: \{lhj20, zqry20, yly20\}@mails.tsinghua.edu.cn).
    
    }
}

\maketitle

\begin{abstract}
Deep hashing has shown promising performance in large-scale image retrieval. However, latent codes extracted by \textbf{D}eep \textbf{N}eural \textbf{N}etworks (DNNs) will inevitably lose semantic information during the binarization process, which damages the retrieval accuracy and makes it challenging. Although many existing approaches perform regularization to alleviate quantization errors, we figure out an incompatible conflict between metric learning and quantization learning. The metric loss penalizes the inter-class distances to push different classes unconstrained far away. Worse still, it tends to map the latent code deviate from ideal binarization point and generate severe ambiguity in the binarization process. Based on the minimum distance of the binary linear code, we creatively propose \textbf{H}ashing-guided \textbf{H}inge \textbf{F}unction (HHF) to avoid such conflict. In detail, the carefully-designed inflection point, which relies on the hash bit length and category numbers, is explicitly adopted to balance the metric term and quantization term. Such a modification prevents the network from falling into local metric optimal minima in deep hashing. Extensive experiments in CIFAR-10, CIFAR-100, ImageNet, and MS-COCO show that HHF consistently outperforms existing techniques, and is robust and flexible to transplant into other methods. Code is available at \href{https://github.com/JerryXu0129/HHF}{https://github.com/JerryXu0129/HHF}.
\end{abstract}


\section{Introduction}

\IEEEPARstart{L}{arge-scale} image retrieval has reached glary attention in the field of computer vision and multimedia computing over the past years~\cite{IR, IR2, ITQ-CCA, KSH, BRE, SDH}. With the arrival of the era of big data, billions of images are uploaded to social platforms and search engines day and night. 
Therefore, how to effectively retrieve approximate nearest neighbors of given queries from massive image databases is a very meaningful research topic with broad applications.

\begin{figure}[ht!]
  \centering
  \includegraphics[width=1\linewidth]{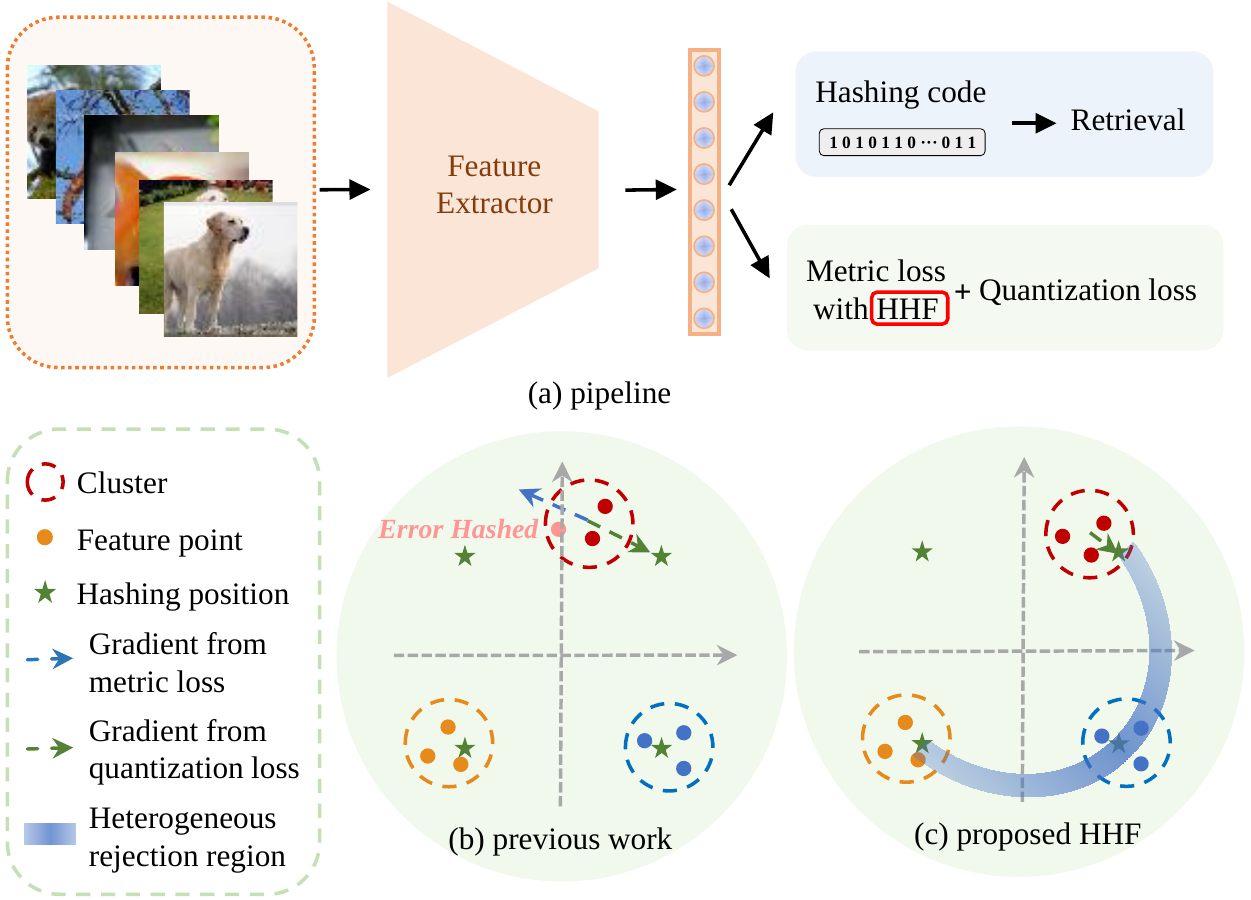}
  \vspace{-18pt}
  	\caption{(a): Standard pipeline for image retrieval and corresponding loss functions with the proposed HHF. (b): the red cluster is balanced under the opposite gradient direction from metric loss and the quantization loss. The possibility of samples being error hashed in binarization increases. (c) the red cluster is only affected by the quantization loss until it reaches the hashing position with the help of HHF.}
  	\label{Fig.intro}
  	\vspace{-20pt}
\end{figure}

Image retrieval, especially content-based retrieval, adopts hashing~\cite{review, deep, hashing} to handle high-dimensional and large-scale data for efficient retrieval and low storage overhead. In light of the promising performance of Deep Neural Networks (DNNs) in feature extraction, Deep Hashing~\cite{CNNH, HashNet, DHN, Proxy-NCA} has become an important research branch in recent years, which aims at mapping high-dimensional feature vectors extracted by DNNs into low-dimensional vectors through specific hash functions. The encoded binary hash codes enable to minimize the storage overhead and maintain the distinguishable feature of given 2D images~\cite{CNNH, HashNet, DHN, Proxy-NCA}.

One of the challenges in Deep Hashing lies in semantic information disappearance in the binarization process (i.e., from continuous latent codes to discrete hash codes). Numerous researchers have carried out in-depth efforts to tackle inevitable information loss in binarization. HashNet~\cite{HashNet} is proposed to tackle information loss by a novel activation strategy. CSQ~\cite{CSQ}, OrthoHash~\cite{OrthoHash} and DCSH~\cite{DCSH} constrain the samples to be aligned with fixed or learnable hash centers to alleviate the information loss. DHN~\cite{DHN} firstly proposed quantization loss during training and demonstrated its flexibility in other works~\cite{DPSH, DSDH, DTSH, DCH, DCWH, IDCH, WSDHQ} by achieving promising performance gains.

However, we figure out an incompatible conflict between metric loss and quantization loss of previous works~\cite{DHN, DTSH, Proxy_anchor_loss}. As Fig.~\ref{Fig.intro}(b) illustrates, the metric loss enforces samples of different classes to be alienated and cluster the same categories. The gradient direction from the metric loss for the red cluster is opposite to that from the quantization loss. Such universal conflict causes the clusters to deviate from the ideal hashing position when converged, which results in error hashed issues and harms retrieval results.

To tackle the above challenges without sacrificing metric learning, the elaborate-designed \textbf{H}ashing-guided \textbf{H}inge \textbf{F}unction (HHF) is proposed to overcome such a dilemma between metric learning and quantization learning. In specific, we propose a novel threshold $\zeta$, determined by hash bit length and category numbers, to prevent inter-distance among different classes (or proxies) being unlimitedly alienated (see Fig.~\ref{Fig.intro}(c)). Inspired by the minimum distance of the binary linear code~\cite{codetable}, we deduce the proper $\zeta$ in different cases and generate a table look-at. The proposed hinge threshold controls negative samples to be punished iff their inter-distance is less than $\zeta$, and finally, keep stabilizing at the hashing position. 

We improve the linear correlated cosine similarity in na\"ive metric loss with the proposed HHF that contains: a). an inflection point $\zeta$ to prevent negative samples unlimited alienated, and b). a relaxation margin $\delta$ to prevent simply mapping similar samples to the same latent codes. We comprehensively validate that the proposed HHF can be easily plugged-and-play and consistently improve the performance among various methods, which shows great superiority and robustness in CIFAR-10, CIFAR-100, ImageNet, and MS-COCO datasets. Our contributions are summarized as follows:

\begin{itemize}
    \item We pinpoint the conflict between metric loss and quantization loss, which is proved harmful for models to converge at the optimal hashing solution. Such a conflict will severely increase error hashed risks that lead to wrong retrieval results.
    \item We overcome such a universal conflict by proposing HHF to modify the metric loss term. HHF is robust in various hash bit lengths and category numbers. We verify the flexibility of HHF with plenty of methods. 
    The proposed HHF achieves better retrieval results and reaches the ideal hash position without deteriorating metric learning.
    \item We conduct extensive experiments to demonstrate that HHF outperforms existing techniques and achieves remarkable state-of-the-art performance gains on four public benchmarks. We will release our code for research purposes in the image retrieval field.
\end{itemize}

The rest of this paper is organized as follows: Section~\ref{sec:relatedwork} briefly reviews the related work. Section~\ref{sec:method} describes the proposed Hashing-guided Hinge Function, which solves the contradiction between metric learning and quantization learning. Section~\ref{sec:exp} demonstrates the effectiveness and flexibility of the proposed method by extensive experiments on four benchmarks, and finally, Section~\ref{sec:conclusion} concludes our work.

\section{Related Work} \label{sec:relatedwork}

Deep Neural Network (DNN), especially Convolution Neural Network (CNN), has shown great superiority in feature extraction of 2D images~\cite{DCNN}. Deep Hashing~\cite{CNNH, DNNH} has surpassed traditional hashing methods~\cite{IR, IR2, ITQ-CCA, BRE, KSH, SDH} and becomes the mainstream for image retrieval. In large-scale image retrieval, Deep Hashing related methods are mainly composed of two parts: 1). metric learning and 2). binary hashing. Therefore, according to different concerns, previous work can be divided into two categories: methods focusing on 1). metric learning and 2). binary hashing. Below we review the methods that are most closely related to our work, and a full in-depth review can be found in~\cite{review, deep, hashing}.

\textbf{Methods focusing on metric learning} can be classified into the pair-based and proxy-based methods. Contrastive Loss~\cite{contrastive, contrastive2} and Pair-wise Loss~\cite{DPSH, HashNet, DSDH, DHN} are primary pair-based methods that mine data-to-data relations from the sample pairs by metric learning. 
However, Pair-wise Loss only correlates given pairs but ignores the relationship between other categories. Hence it will easily drop into locally optimal solutions and cannot deal with hard pair challenges.

To tackle the challenge, Triplet Loss~\cite{Triplet, DNNH, DTSH} associates the anchor input with a positive and a negative sample. It tackles the hard triplet by comparing the distance between the anchor and the two samples. Based on Triplet Loss, Sohn et al.~\cite{N-pair} propose $N$-pair Loss that contains an efficient batch construction strategy to optimize one positive sample and $N-1$ negative samples simultaneously, which speeds up convergence and reduces computational complexity. Lifted Structure Loss~\cite{Lifted} also considers the distance among one positive anchor and multiple negative anchors, and penalizes them according to their hardness, alleviating the challenging hard pair problem. Although pair-based losses~\cite{DPSH, Triplet} can sufficiently exploit the relationships of given samples and accurately model the distance of each category, the large computational overhead limits these methods to consider the whole datasets but only constrains on a single mini-batch. As a result, metric learning based on pair-based methods only contains limited local information.

Proxy-based methods are proposed to alleviate the computation complexity of pair-based losses.
Proxy-NCA~\cite{Proxy-NCA} replaces the pair-wise error with a set of learnable class centers (i.e., proxies), and calculates the distance between each proxy and positive/negative samples. Because the category number is far less than samples, Proxy-based losses significantly improve the convergence speed by only considering the distance of each sample and proxy.
Combining the advantages of proxy and $N$-pair Loss, Manifold Proxy Loss~\cite{Manifold} utilizes the manifold-aware distance as the metric in the feature space to improve performance further. Proxy-Anchor Loss~\cite{Proxy_anchor_loss} serves as the state-of-the-art proxy-based method for metric learning. It integrates both advantages of the two methods that adopts \textit{LogSumExp} function to excavate semantic relations of different samples and solves hard sample challenges effectively.

\textbf{Methods focusing on binary hashing} aim at dealing with information loss. The latent codes obtained by DNNs will be hashed into discrete $\pm 1$ for easy storage and fast retrieval. However, plenty of continuous semantic information will get lost during the binarization process. To tackle the problem, HashNet~\cite{HashNet} optimizes the pair-wise loss with non-smooth binary activations, and addresses the gradient difficulty in optimization via \textit{sign} activation function $\text{sgn}(\cdot)$. The gradient of \textit{Tanh} is close to $0$ in the domain, while the considerable distance between the latent codes of similar samples would lead to a minor penalty, hence it cannot guarantee all similar samples well clustered in the feature space.

In addition, some methods directly learn the semantic relationship between samples and the predefined hash centers.
CSQ~\cite{CSQ} constructs fixed hash center sets randomly or based on the Hadamard matrix, which alleviates information loss during the binarization process. Based on the orthogonal hashing centers~\cite{CSQ}, OrthoHash~\cite{OrthoHash} optimizes the distance between samples and hash centers by proposing Scaled Cosine Similarity loss. However, the artificially defined hash centers miss the semantic relationship of intra-class and inter-class. The Hadamard matrix-based or randomly generated codes are either inflexible to deal with the changed length and category number or cannot achieve satisfactory performance.

Compared to those hash center fixed methods~\cite{CSQ,OrthoHash}, DCWH~\cite{DCWH} and IDCH~\cite{IDCH} set learnable hash centers and optimize them by calculating the sample offset and voting per bit in a mini-batch. DCSH~\cite{DCSH} obtains the latent codes in each epoch and calculates the average of each bit to obtain the binarized hash centers.
Pair-wise loss based DHN~\cite{DHN} integrates quantization loss with metric loss to constrain latent codes, and controls the quantization error to improve hashing quality. Most recent works~\cite{DPSH, DSDH, DTSH, CSQ, DCWH, IDCH, WSDHQ} add the quantization loss to reduce information loss during binarization and show promising performance gains. 

However, we figure out an incompatible conflict when combining metric loss with quantization loss, which causes much information loss during the binarization process. Hence, the carefully designed HHF overcomes such contradiction, which can be applied in most existing techniques and achieve significant performance gains.

\section{Methodology} \label{sec:method}

\subsection{Problem Definition}

Given $M$ images $\mathcal{X} = \{x_i\}_{i=1}^M$ and corresponding label matrix $\mathcal{Y} \in \mathds{R}^{C\times M}$, where $C$ is the category numbers of the whole dataset, Deep Hashing learns a feature extractor $\mathcal{E}$ parameterized by $\Theta$ and a hashing function $\mathcal{H}$. For each $x_i \in \mathcal{X}$, we first obtain the latent code $\bm{h_i}=\{h_1h_2\cdots h_K|h_i\in\mathbb{R}\}\in \mathbb{R}^K$ by $\mathcal{E}$ with the supervision of a series of losses: $\bm{h_i} \leftarrow  \mathcal{E}(x_i|\Theta)$. Since we adopt the binarization as hashing function $\mathcal{H}$, the latent code $\bm{h_i}$ is quantizated into a compact $K$-bit binary code $\bm{b_i} = \{b_1b_2\cdots b_K|b_i\in\{\pm 1\}\}$ by the hashing function $\mathcal{H}$: $\bm{b_i} \leftarrow \mathcal{H}(\bm{h_i})$. We reasonably assume that the hashing code $\bm{b}$ in the Hamming space shares the similar semantic information with the representation $\bm{h_i}$ in the feature space. For given query image $x_q$, we sort the hash codes for all the samples in the database according to their Hamming distance, and return the top-$N$ images as the query results. The core challenge of this task is to learn a reliable feature extractor $\mathcal{E}$ to cluster images of different categories with proper and distinguishable hash positions.

\subsection{Loss Functions in Image Retrieval}
\textbf{Metric Loss.}
Without loss of generality, for any two $K$-dimension binary code $\bm{b_1},\bm{b_2} \in \{\pm 1\}^K$, the Hamming distance estimates the similarity of the two binary codes, which is defined as the summary of \textit{XOR} operation per bits as Eq.~\ref{Eq.hamming}.
\begin{equation}
    \begin{aligned}
    \mathbb{D}_{\text{Ham}}(\bm{b_1},\bm{b_2}) &= \sum_{i = 1}^{K}\mathds{1}({b_1}_i \neq {b_2}_i)
    \end{aligned}
    \label{Eq.hamming}
\end{equation}

However, in practice, the Hamming distance in Eq.~\ref{Eq.hamming} is non-differentiable, which is not qualified to measure the distance among latent codes for training. Considering the Hamming distance $\mathds{D}_{\text{Ham}}(\bm{b_1},\bm{b_2})$ is correlated to cosine similarity $\cos(\bm{b_1},\bm{b_2})$, cosine similarity is adopted instead to approximate the distance between latent code pairs $\bm{h_1},\bm{h_2}$ for backpropagation. i.e., the distance between them only depends on their angle in hypersphere space as Eq.~\ref{Eq.cosine}.
\begin{equation}
    \begin{aligned}
    \mathbb{D}_{\text{Ham}}(\text{sgn}(\bm{h_1}),&\text{sgn}(\bm{h_2})) =\frac{K}{2}\left (1-\frac{\bm{b_1}\cdot \bm{b_2}}{||\bm{b_1}||\cdot ||\bm{b_2}||} \right) \\
    &\propto -\cos(\bm{b_1},\bm{b_2}) \approx  -\cos(\bm{h_1},\bm{h_2})
    \end{aligned}
    \label{Eq.cosine}
\end{equation}

Metric loss aims at clustering, i.e., to push different category samples far as possible and pull samples of the same categories close as possible. In specific, existing techniques~\cite{DPSH, DSDH, DTSH, CSQ, DCWH, IDCH, WSDHQ} generally adopt the estimated cosine similarity in a mini-batch $\mathcal{B}$ to design such loss term, which can be abstractly recorded as Eq.~\ref{Eq.metricloss}:
\begin{equation}
    \label{Eq.metricloss}
    \begin{aligned}
    \mathcal{L}_{\text{Metric}} &=\sum_{\bm{h_1} \in \mathds{H}}\sum_{\substack{\bm{h_2} \in \mathds{H}'}} \mathds{1}(y_1=y_2)\mathcal{F}_+\left(-\cos(\bm{h_1}, \bm{h_2})\right) \\
    &\quad + \sum_{\bm{h_1} \in \mathds{H}}\sum_{\substack{\bm{h_2} \in \mathds{H}'}} \mathds{1}(y_1\neq y_2)\mathcal{F}_-\left(\cos(\bm{h_1}, \bm{h_2})\right),
    \end{aligned}
\end{equation}
where $\mathds{1}(\cdot)$ is the indicator function that equals $1$ iff $(\cdot)$ is true, otherwise is $0$. $\mathcal{F}_+$ and $\mathcal{F}_-$ indicate the designed loss functions for positive or negative samples, respectively, which either estimate the similarity of pair-based methods or proxy-based methods. $\mathds{H}$ represent the latent codes of given mini-batch images, $\mathds{H}'$ indicates the set of other samples' latent codes or class proxies. $y_i$ indicates the class label.

\textbf{Quantization Loss.}
However, in the image retrieval field, metric loss considers neither efficient query nor easy storage requirements, but it only focuses on clustering images (or latent codes) of the same categories and alienating those of different labels. Therefore, metric loss cannot guarantee the binarized latent codes well-located in the ideal hash position because of the missing quantization constraints. If the latent code $\bm{h}$ is directly binarized, a large amount of feature information will get lost. e.g., some bits close to $0$ of similar samples will be ambiguously binarized to $\pm 1$, which is highly undesirable for clustering images with similar hash codes.

To handle ambiguity issues that cause information loss, quantization loss~\cite{DHN} is proposed to constrain latent codes. By adding quantization loss~\cite{DSDH, DTSH, CSQ, DCWH, IDCH, WSDHQ}, retrieval performance is demonstrated to achieve promising improvement. It's often calculated by $l_1$ or $l_2$ regulation loss between latent codes $\bm{h}$ and their corresponding binary vectors $\bm{b}$ obtained by sign function $\text{sgn}(\cdot)$:
\begin{equation}
\label{Eq.quan} 
   \mathcal{L}_{\text{Quan}} = \sum_{\bm{h}\in \mathds{H}}||\bm{h} - \text{sgn}(\bm{h})||_F^n,
\end{equation}
where existing techniques~\cite{DPSH, DSDH, DTSH, CSQ, DCWH, IDCH, WSDHQ} generally set $n=1,2$ to compute $l_1$ or $l_2$ loss of the quantization term.

\begin{figure*}[htp!]
  \centering
  \includegraphics[width=1\linewidth]{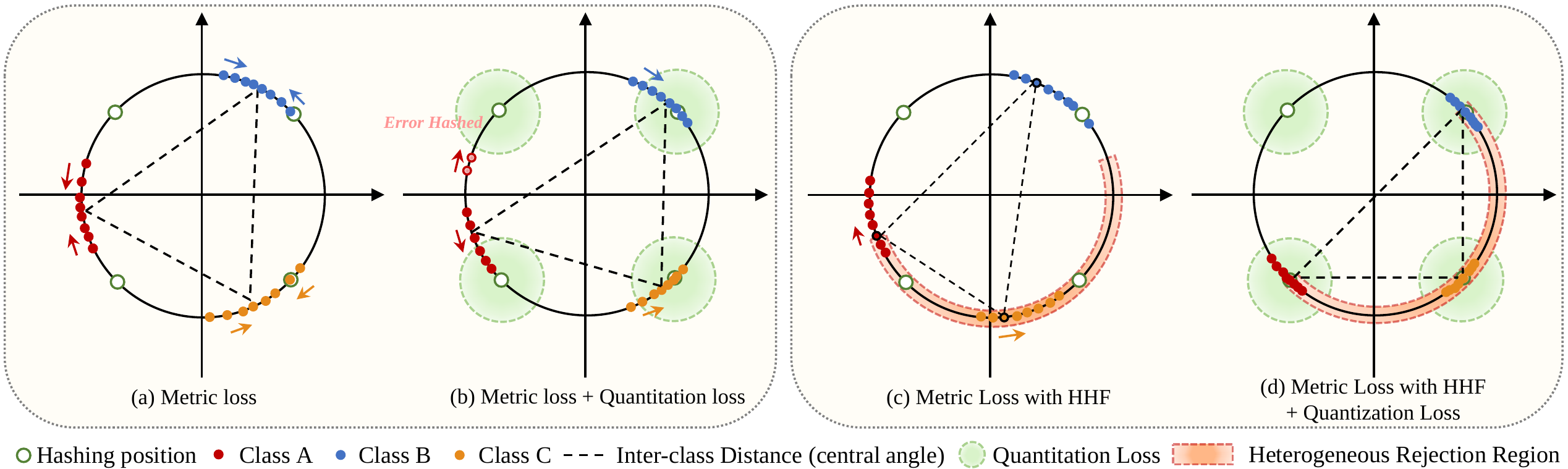}
  \vspace{-15pt}
   	\caption{Comparison between metric loss with/without HHF. (a) Metric loss pushes different category samples far as possible in the feature space. (b) After adding the quantization loss, the class centers will move towards the binary points. But due to the repulsive effect of other classes in the metric loss, the class centers cannot fit the hashing position. (c) Our HHF method only encourages sample (or proxy) to push away samples in different categories whose angle with the sample (or proxy) is $\le 90^\circ$. (d)  After adding quantization loss to metric loss with HHF, the result achieves the global optimal solution of this case.}
    \label{Fig.motivation}
    \vspace{-15pt}
\end{figure*}

\subsection{Conflict on Metric Loss and Quantization Loss}
However, the learning purposes of metric loss and quantization loss are inherently contradictory. We intuitively illustrate such a conflict in Fig.~\ref{Fig.motivation} for clarity, which contains three different categories in a 2D feature space (i.e., a hypersphere).

Therefore, all samples in Fig.~\ref{Fig.motivation} are scaled to a feature circle with radius $ r = \sqrt{2}$ in 2D space (correspondingly, in $K$-dimension space, all samples are distributed on a hypersphere with radius $ r = \sqrt{K}$). The binarization operation directly pulls all samples to four binary points (i.e., $(\pm 1,\pm 1)$), corresponding to four quadrants. The more samples away from the binary points, the more information will be lost after hashing $\mathcal{H}$.

In Fig.~\ref{Fig.motivation}(a), the motivation of metric learning is to push the clustering centers of different categories as far as possible. Then, finally, it forms an equilateral triangle on the 2D space to obtain the optimal solution (i.e., each cluster is pushed far as possible). At the same time, the purpose of quantization learning in Fig.~\ref{Fig.motivation}(b) is to pull the class centers to one of the four binary points as close as possible to get the optimal solution for quantization (i.e., towards least information loss in the binarization process). 

It is intuitive to find the incompatible conflict where the optimal metric and quantization solutions cannot be satisfied simultaneously. In fact, the ideal situation is that the three class centers are strictly at three of the four binary points, which is the optimal global solution (Fig.~\ref{Fig.motivation}(d)) that limited information lost during the binarization operation.

Methods integrate metric loss and quantization loss will finally get an intermediate situation (Fig.~\ref{Fig.motivation}(b)) between the optimal metric solution (equilateral triangle) and the optimal global solution (isosceles right triangle), resulting in more information disappeared in the binarization process with poor retrieval accuracy. Specifically, samples farther from the clustering centers are more likely to be misclassified to other binary points (as we will prove in Fig.~\ref{Fig:retrieval1}).

\subsection{HHF: Novel Conflict Solution}

\begin{figure}[htp!]
  \centering
  \includegraphics[width=1\linewidth]{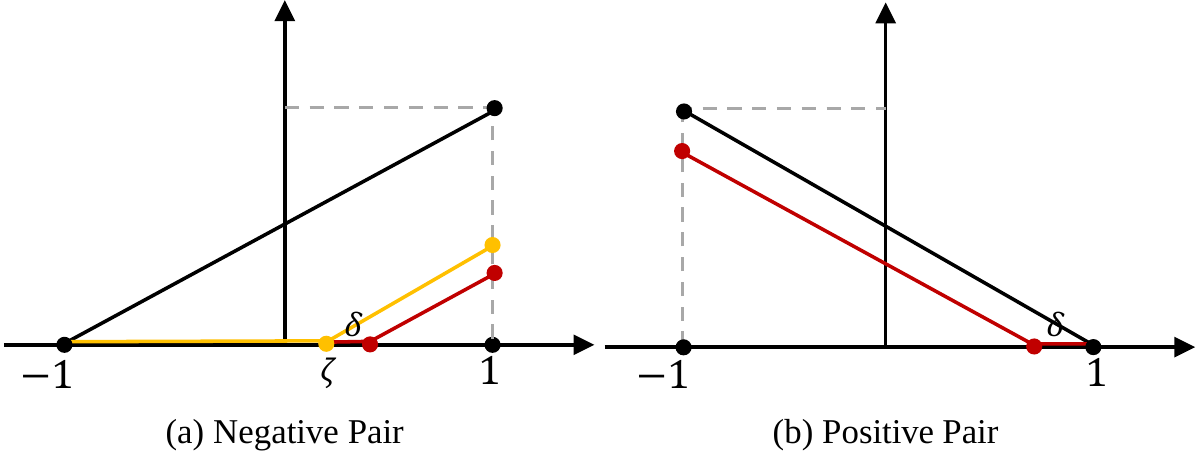}
  \vspace{-15pt}
   	\caption{Yellow line in (a) shows that we modify the linear correlation of cosine similarity in metric loss with a hinge function by a specific inflection point $\zeta$ for negative pairs. Red lines in (a)(b) illustrate the hinge function after optimization for both negative and positive pairs.}
   	\label{Fig.cos_sim}
   	\vspace{-15pt}
\end{figure}

\textbf{Overview of HHF.} We propose Hashing Guided Hinge Function (HHF) to solve the conflict between metric loss and quantization loss. The effectiveness of HHF is intuitively illustrated in Fig.~\ref{Fig.motivation}(c)(d). We modify cosine similarity in metric loss with hinge function controlled by a specific inflection point $\zeta$, shown as the yellow line in Fig.~\ref{Fig.cos_sim}(a). The inflection point $\zeta$, which depends on code length and category number, illustrates the minimum distance of the optimal global solution $d_{\min}$. We do not expect the metric distances between different classes to be infinitely alienated. Instead, we enforce to push them apart until reaching a certain distance $\zeta$, deduced by $d_{\min}$. When the distance between different classes satisfies the $\zeta$, we consider the optimal metric solution is achieved.

In Fig.~\ref{Fig.motivation}, when the cosine similarity between two categories is exactly $0$, the optimal global solution can be satisfied. For any samples (or class proxies), HHF only pushes away those different category samples until their angle reaches $90^\circ$, and does not constrain angles greater than $90^\circ$. In this way, the optimal metric solution does not conflict with the optimal quantization solution, and finally, the optimal global solution can be obtained.

Noteworthy, inspired by the Triplet Loss~\cite{Triplet}, our empirical study shows that models will be overfitting if all positive samples are uncontrolled close to each other. Therefore, we make an additional margin on intra-class and inter-class distance. To be specific, we add a hyperparameter $\delta$ to the original HHF as the relaxation factor. The optimized cosine similarity function is shown as the red line in Fig.~\ref{Fig.cos_sim}(a)(b).

\textbf{Details of Inflection Constant $\bm{\zeta}$.}
We then discuss the value and calculation approach of obtaining the inflection point constant $\zeta$. The target is equivalent to minimizing the distance of given binary linear codes. However, it proves to be an NP-hard problem~\cite{code}, i.e., calculating the ideal minimum distance requires exponential time complexity. Fortunately, researchers~\cite{codetable} propose to estimate the upper and lower bound of the minimum distance, which constrains the minimum distance through serious inequalities and presents minimum distance bound tables $\mathcal{T}$ for binary linear code.

Inspired by~\cite{codetable}, the minimum distance $d_{\min}$ is set according to the following rules. a). If the value obtained from $\mathcal{T}$ is a constant, then we take this value as $d_{\min}$. b). Otherwise, if the value obtained is a range, then the median of this range is chosen as $d_{\min}$. Finally, to calculate the ideal inflection point for metric learning, we obtain the value of $\zeta$ by normalizing $d_{\min}$ according to the code length, i.e., numerically equivalent to the hash code $\bm{b}$'s length $K$:
\begin{equation}
    \begin{aligned}
    \zeta = 1 - 2 \times \frac{d_{\min}}{K}
    \end{aligned}
    \label{Eq.zeta}
\end{equation}

For convenience, we generate a $256\times 256$ table that can directly obtain the value of the inflection point $\zeta$ uniquely determined by the hash code length and category numbers, which is entirely available in our public code repository.

\textbf{Proxy-Anchor Loss with HHF.}
To better quantify each input $x\in \mathcal{X}$ into an appropriate hash code, a proper loss function is required to supervise the metric distance. We choose the previous state-of-the-art approach Proxy-Anchor~\cite{Proxy_anchor_loss} as the backbone metric on HHF for illustration. The metric loss of Proxy-Anchor is calculated as Eq.~\ref{Eq.Proxy-Anchor}:
\begin{equation}
\label{Eq.Proxy-Anchor}
    \begin{aligned}
    \mathcal{L} &=\frac{1}{|\mathds{P}|} \sum_{\bm{p} \in \mathds{P}} \log \left( 1+ \sum_{\bm{h} \in \bm{h_p^-}} e^{\alpha(\cos(\bm{h},\bm{p})+\gamma)} \right) \\
    &+\frac{1}{|\mathds{P}^+|} \sum_{\bm{p} \in \mathds{P}^+} \log \left( 1+ \sum_{\bm{h} \in \bm{h_p^+}} e^{-\alpha(\cos(\bm{h},\bm{p})-\gamma)}\right),
    \end{aligned}
\end{equation}
where $\alpha$ is a scaling factor, and $\gamma$ is a margin. $\mathds{P}$ denotes the set of the proxies and $\mathds{P}^+$ represents the positive proxies of samples in a mini-batch $\mathcal{B}$. For a proxy, the set of latent codes $\mathds{H} =\{\bm{h_i}\}_{i=1}^{\mathcal{B}}$ is categorized into two sets: $\bm{h^+_p}$ indicates the set of all samples of same categories, and the rest $\bm{h^-_p}$ = $\mathds{H} - \bm{h^+_p}$ indicates different classes to proxy. 

In Eq.~\ref{Eq.Proxy-Anchor}, the error component $-\alpha(\cos(\bm{h},\bm{p})-\gamma)$ and $\alpha(\cos(\bm{h},\bm{p})+\gamma$) is linear correlated to cosine similarity for all samples, which contradicts quantization learning as we analyzed before. According to the proposed HHF, we modify the linear calculation part of cosine similarity in na\"ive Proxy-Anchor Loss to the Hinge function with a specific inflection point $\zeta$, as Eq.~\ref{Eq.Metric} illustrates:
\begin{equation}
\label{Eq.Metric}
    \begin{aligned}
    \mathcal{L}&_{\text{Metric}} =\frac{1}{|\mathds{P}|} \sum_{\bm{p} \in \mathds{P}} \log \left( 1+ \sum_{\bm{h} \in \bm{h_p^-}} (e^{\alpha H(\cos(\bm{h},\bm{p}), -\zeta - \delta)} - 1) \right)\\
    & +\frac{1}{|\mathds{P}^+|} \sum_{\bm{p} \in \mathds{P}^+} \log \left( 1+ \sum_{\bm{h} \in \bm{h_p^+}} (e^{\alpha H(-\cos(\bm{h},\bm{p}), 1 - \delta)} -1) \right),
    \end{aligned}
\end{equation}
where $H(x, y) = \max(0, x + y)$ is a hinge loss function, $\delta$ is a relaxation factor to prevent overfitting. Since the cosine similarity with HHF is always $\ge 0$, its exponential part is $\ge 1$. As a result, each sample has limited influence on loss when calculating gradient, which damages backpropagation, especially when the large batch size is adopted. Therefore, we set the exponential part minus to $1$ to enhance the influence of each sample and enlarge the gradient.

In addition, Proxy-Anchor only considers metric loss but ignores the quantization term, hence cannot perform as expected for retrieval. Therefore, we follow~\cite{DTSH, IDCH} to adopt $l_2$ regularization as quantization loss $\mathcal{L}_{\text{Quan}}$, so as to extend Proxy-Anchor Loss to obtain the final modified loss function that integrates the metric term $\mathcal{L}_{\text{Metric}}$ and the quantization term $\mathcal{L}_{\text{Quan}}$ by weight $\beta$ as Eq.~\ref{Eq.total_loss}.
\begin{equation}
\label{Eq.total_loss}
   \mathcal{L}_{\text{Total}} = \mathcal{L}_{\text{Metric}} + \beta \mathcal{L}_{\text{Quan}}
\end{equation}

\textbf{Learning Algorithm with HHF.}
HHF is flexible and robust to plugged-and-play into other methods that require metric terms and quantization terms. Algo.~\ref{Alg:training_manner} shows the training algorithm of given proxy-based loss as an illustration. The algorithm of pair-based losses with HHF is in a similar fashion.

\begin{algorithm}[ht!]
    \caption{Training process with the proposed HHF.}
    \label{Alg:training_manner}
	\renewcommand{\algorithmicrequire}{\textbf{Input:}}
	\renewcommand{\algorithmicensure}{\textbf{Output:}}
	\begin{algorithmic}[1]
		\REQUIRE Training images $\mathcal{X}=\{x_i\}^M_{i=1}$, label matrix $\mathcal{Y}\in \mathds{R}^{C\times M}$, hash code length $K$, mini-batch $\mathcal{B}$.
		\ENSURE Optimized model param $\Theta^*$, encoded hash position $\mathds{B}^*\in \{\bm{b}\}^M_{i=1}$, proxy matrix $\mathds{P}^*\in \mathds{R}^{C\times N}$.
		\STATE Calculate the dimension $k$ of the corresponding binary linear code by the number of $C$ categories: $k= \lceil \log C \rceil$;
		\STATE Look-up the Minimum Distance Bound $\Omega= [\Omega_{\min},\Omega_{\max}]$ of $k$-dim $K$-bits binary linear code via $\mathcal{T}$~\cite{codetable} and obtain the $d_{\min}$ according to our rule;
		\STATE Normalize $d_{\min}$ according to Eq.~\ref{Eq.zeta} to obtain the ideal $\zeta$;
		\STATE Random initialize $\Theta \leftarrow \Theta^{(0)}$, $\bm{\mathds{P}} \leftarrow \bm{\mathds{P}}^{(0)}$;
		\STATE Initialize Training epoch $T\leftarrow 0$;
		\REPEAT
	    \STATE Randomly sample a mini-batch data $\mathcal{X}'=\{x_i\}^{\mathcal{B}}_{i=1}$;
		\STATE Compute 1). latent codes $\bm{h_i} \leftarrow  \mathcal{E}(x_i|\Theta)$ and 2). hash codes $\bm{b_i} \leftarrow \text{sgn}(\bm{h_i})$ by forward propagation;
    		\FOR{$\bm{h}=\bm{h_1}$ to $\bm{h_\mathcal{B}}$, $\bm{p}=\bm{p_1}$ to $\bm{p_C}$} 
            		\IF{$\bm{h}$ and $\bm{p}$ is a Positive Pair}
                	    \STATE the similarity is $\max(0, 1-\cos(\bm{h},\bm{p})-\delta)$
                	\ELSIF{$\bm{h}$ and $\bm{p}$ is a Negative Pair}
                	    \STATE the similarity is $\max(0, \cos(\bm{h},\bm{p})-\zeta-\delta)$
                	\ENDIF
                	\STATE Compute Metric loss $\mathcal{L}_{\text{Metric}}$ via Eq.~\ref{Eq.metricloss};
    	    \ENDFOR
    	    \STATE Compute Quantization loss $\mathcal{L}_{\text{Quan}}$ via Eq.~\ref{Eq.quan} and total loss  $\mathcal{L}_{\text{Total}}$ via Eq.~\ref{Eq.total_loss};
	    \STATE Update $\Theta^{(T+1)}$ and $\mathds{P}^{(T+1)}$ by back propagation;
	    \STATE $T\leftarrow T+1$;
    	\UNTIL{Convergence}
    	\STATE Return $\Theta^{*} \leftarrow \Theta^{(T)}, \mathds{B}^*\leftarrow \mathcal{E}(\mathcal{X}|\Theta^*), \mathds{P}^{*} \leftarrow \mathds{P}^{(T)}$.
	\end{algorithmic}
\end{algorithm}

\section{Experiments} \label{sec:exp}

In this section, we verify the effectiveness of the proposed HHF by conducting extensive experiments on four representative benchmarks (w.r.t. CIFAR-10~\cite{cifar}, CIFAR-100~\cite{cifar}, MS-COCO~\cite{COCO}, and ImageNet~\cite{ImageNet}) for comprehensive comparisons. We integrate HHF with various off-the-shelf metric losses~\cite{Proxy_anchor_loss,Proxy-NCA,DHN} and compare them to the current state-of-the-art methods quantitatively and qualitatively. In addition, we conduct hyperparameter sensitivity experiments to fully demonstrate the robustness of the proposed method.

\vspace{-8pt}

\subsection{Datasets}
\textbf{CIFAR-10}~\cite{cifar} is a widely used dataset for image recognition and is appropriate for image retrieval tasks. It contains $60,000$ images at $32\times32$ resolution with $10$ different categories.
We follow previous work~\cite{DPSH, DTSH, DSDH} to construct CIFAR-10-MINI and CIFAR-10-FULL with different protocols for experiments. In CIFAR-10-MINI, we randomly select $500$ and $100$ images from each class as the train set and query set, respectively, the remaining images as the database. In CIFAR-10-FULL, $50,000$ images both serve as the train set and database, and the rest $10,000$ images as the query set.

\textbf{CIFAR-100}~\cite{cifar} contains the same image numbers ($60,000$ in total) as CIFAR-10, while it has $100$ classes with $600$ images per category. Follow~\cite{DCWH, IDCH}, we use the official $50,000$ images as train set and database, and $10,000$ images for query in our experiment.

\textbf{MS-COCO}~\cite{COCO} is a popular multi-label dataset for image recognition. It contains more than $120,000$ images, each of which is labeled as several of $80$ categories. Follow previous work~\cite{CSQ, DCWH, IDCH}, we first preprocess to remove the images without any labels from the dataset and then randomly select $5,000$ images from the remaining $122,218$ ones, containing at least one label, as the query set, $10,000$ images as the training set, and the rest as the database.

\textbf{ImageNet}~\cite{ImageNet} is one of the most popular benchmark datasets for the Large Scale Visual Recognition Challenge (ILSVRC 2012). It contains more than $120,000$ images in $1,000$ categories. We use the same data and settings as~\cite{HashNet, DCWH}. The main challenge lies in training sample paucity and the large number of categories.

\vspace{-8pt}

\subsection{Implementation Details}
The proposed HHF is implemented in the PyTorch framework~\cite{pytorch}. We comprehensively adopt ResNet-50~\cite{ResNet} and GoogLeNet~\cite{googlenet} pretrained on ImageNet to illustrate the robustness of our HHF. We fine-tune the pre-trained backbones for all layers up to the FC layer and map the output layer dimension to the hash bit $K$. We adopt stochastic gradient descent (SGD)~\cite{SGD} as the optimizer with momentum $0.9$ and weight decay $5e-4$. The initial learning rate is $0.001$ for feature extraction and $0.01$ for proxy adjustment if the proxy-based loss is adopted as the baseline metric. The learning rate decreases by $0.5$ every $10$ epochs for $100$ epochs in total. For the hyperparameters, if without special instructions, $\delta$ in the HHF method is empirically set to $0.2$, and for the rest of the hyperparameters related to other methods, we follow the settings in the previous work~\cite{DHN, Proxy-NCA, Proxy_anchor_loss}.

\subsection{Evaluation Metrics}
Given any query image, we retrieve top-$N$ images in the database based on the closest Hamming distance and calculate their similarity. We adopt the standard similarity protocol from the previous work~\cite{HashNet, DPSH, DCWH}. For single-label datasets (i.e., CIFAR, ImageNet), two images are considered similar iff they have the same label. For multi-label datasets (i.e., MS-COCO), two images are considered similar iff they share at least one label. We evaluate the proposed method with several strong baselines~\cite{CNNH, DNNH, DHN, DPSH, DTSH, DSDH, HashNet, Proxy-NCA, DCH, DCCH} and current state-of-the-art methods~\cite{DCWH, CSQ, IDCH, Proxy_anchor_loss, DCSH, OrthoHash} for comprehensive comparisons. 

\textbf{mean Average Precision} (mAP) serves as the primary metric in our experiment, which is the most widely used metric in image retrieval.
Average Precision (AP) refers to the area under the Precision-recall curve. Larger AP indicates better retrieval performance. mAP is the mean of average precision for each query image that can be calculated as Eq.~\ref{Eq.map}.
\begin{equation}
    \text{mAP}@N = \frac{1}{|\mathcal{Q}|} \sum_{x_q \in \mathcal{Q}} AP(x_q,y_q)@N
    \label{Eq.map}\vspace{-5pt}
\end{equation}
where $N$ is the number of relevant images w.r.t. the query image $x_q \in \mathcal{Q}$ with corresponding label(s) $y_q$.

\textbf{Hash Position Error} (HPE) is proposed and adopted to measure the quality of quantization learning in image retrieval quantitatively. Specifically, HPE measures the average $l_2$ distance between the latent codes and corresponding hash codes, as Eq.~\ref{Eq.hashdist} illustrates.
\begin{equation}
    \text{HPE} =\frac{1}{N} \sum_{i = 1}^{N}||\bm{h_i}-\text{sgn}(\bm{h_i})||_{2}^2,
    \label{Eq.hashdist}\vspace{-5pt}
\end{equation}
where $\text{sgn}(\cdot)$ is a sign function that maps continuous latent codes into binary values. Smaller HPE indicates that the latent codes are closer to the ideal hash position, and shows less information loss during the binarization process.

\textbf{Global Inter-intra Distance Ratio.}
Apart from mAP and HPE, we follow~\cite{IDCH} to measure the Global Inter-intra Distance Ratio $\eta_{\text{global}}$ for quantitative analyses on the quality of metric learning. $\eta_{\text{global}}$ refers to the average distance ratio of intra-class and inter-class, which is defined as Eq.~\ref{Eq.hashdistratio}:
\begin{equation}
    \eta_{\text{global}} = \frac{\frac{1}{N}\sum\limits_{i=1}^N ||\bm{h_i} -\bm{\bar{h}^c_i}||_2^2}{\frac{1}{C}\sum\limits_{i=1}^C \frac{1}{C-1}\sum\limits_{\substack{i\neq j,j=1}}^C ||\bm{\bar{h}^c_i} - \bm{\bar{h}^c_j}||_2^2 },
    \vspace{-6pt}
    \label{Eq.hashdistratio}
\end{equation}
where $\bm{\bar{h}^c_i}$ represents the class center of the $i\_$th class. The Global Inter-intra Distance Ratio considers the average distance among class centers, i.e., $\eta_{\text{global}}$ estimates the inter-intra distance of each sample among all other categories.

\textbf{Local Inter-intra Distance Ratio.}
In practice, we notice those semantic-distinguishable classes (or remote centers), which away from the given class, have seldom retrieval side-effect. Hence we believe the evaluation of the most confusing classes is necessary. To achieve this, we propose Local Inter-intra Distance Ratio $\eta_{\text{local}}$ to estimate the worst entangled class centers, which is defined as Eq.~\ref{Eq.localratio}.
\begin{equation}
    \eta_{\text{local}} = \frac{1}{N}\sum_{i=1}^N\frac{||\bm{h_i} - \bm{\bar{h}^c_i}||_2^2}{||\bm{h_i} - \bm{\bar{h}^c_{\phi(i)}}||_2^2},
    \label{Eq.localratio}
\end{equation}
where $\phi(i)=\mathop{\arg\min}\nolimits_{\substack{i\neq j}}||\bm{h_i} - \bm{\bar{h}^c_j}||_2^2$ represents the closest class center index of the $i\_$th center. Both $\eta_{\text{global}}$ and $\eta_{\text{local}}$ measure the clustering quality of metric learning, smaller $\eta_{\text{global}}$ and $\eta_{\text{local}}$ indicates better performance in clustering.

\vspace{-8pt}
\subsection{Results on CIFAR-10 and CIFAR-100}

\begin{table*}[htp!]
\centering
\normalsize
\caption{mAP performance by Hamming Ranking for different hash bits on CIFAR-10-MINI, CIFAR-10-FULL, and CIFAR-100-FULL. $^\star$: reported results with the same GoogLeNet backbone from \cite{DCWH, IDCH}. $^\dagger$: our reproduced results with GoogLeNet.}
\vspace{-6pt}
\resizebox{1\linewidth}{!}{%
\begin{tabular}{l|cccc|cccc|cccc}
\toprule
Dataset   & \multicolumn{4}{c|}{CIFAR-10-MINI}                              & \multicolumn{4}{c|}{CIFAR-10-FULL}                              & \multicolumn{4}{c}{CIFAR-100-FULL}                                     \\ \midrule
Hash bits & 12 bits        & 24 bits        & 32 bits        & 48 bits        & 12 bits        & 24 bits        & 32 bits        & 48 bits        & 12 bits        & 24 bits        & 32 bits        & 48 bits        \\ \midrule
DPSH$^\star$~\cite{DPSH}    & 0.797          & 0.806          & 0.820          & 0.802          & 0.908          & 0.922          & 0.925          & 0.935          & 0.060          & 0.101          & 0.120          & 0.159          \\
DTSH$^\star$~\cite{DTSH}   & 0.790          & 0.797          & 0.794          & 0.775          & 0.928          & 0.935          & 0.940          & 0.942          & 0.607          & 0.706          & 0.712          & 0.725          \\
DSDH$^\star$~\cite{DSDH}    & 0.800          & 0.802          & 0.804          & 0.808          & 0.913          & 0.925          & 0.943          & 0.930          & 0.078          & 0.150          & 0.187          & 0.227          \\
DCWH$^\star$~\cite{DCWH}     & 0.818          & 0.840          & 0.848          & 0.854          & 0.940          & 0.950          & 0.954          & 0.952          & 0.723          & 0.744          & 0.757          & 0.766          \\
IDCWH$^\star$~\cite{IDCH}    & 0.828          & 0.865          & 0.868          & 0.849          & 0.964          & 0.969          & 0.967          & 0.968          & 0.764          & 0.813          & 0.824          & 0.835          \\
Proxy-Anchor$^\dagger$~\cite{Proxy_anchor_loss}     & 0.878          & 0.886          & 0.889          & 0.904          & 0.975   & 0.976   & 0.977   & 0.978   & 0.773          & 0.860          & 0.870          & 0.870          \\ \midrule
Proxy-Anchor$^\dagger$ + HHF       & \textbf{0.889} & \textbf{0.903} & \textbf{0.909} & \textbf{0.908} & \textbf{0.975} & \textbf{0.976} & \textbf{0.978} & \textbf{0.979} & \textbf{0.851} & \textbf{0.868} & \textbf{0.873} & \textbf{0.875} \\ 
\bottomrule
\end{tabular}}
\vspace{-6pt}
\label{Tab.CIFAR}
\end{table*}

\begin{table*}[htp!]
\centering
\setlength{\tabcolsep}{6pt}
\caption{mAP performance with different hash bits on ImageNet and MS-COCO. $^\star$: reported results from~\cite{HashNet, DCWH, CSQ, IDCH, DCCH, OrthoHash, DCSH}. $^\dagger$: our reproduced results with GoogLeNet and ResNet-50 through the open-source code and the loss equation by existing literature. $\Delta$: average improvement with HHF for different hash bits compared with the original method.}
\vspace{-6pt}
\normalsize
\resizebox{1\linewidth}{!}{%
\begin{tabular}{l|l|ccccc|ccccc}
\toprule
          & Dataset            & \multicolumn{5}{c|}{ImageNet (mAP@1000)}                                     & \multicolumn{5}{c}{MS-COCO (mAP@5000)}                                        \\ \midrule
Backbone  & Hash bits         & 16 bits        & 32 bits        & 48 bits        & 64 bits        &   $\Delta$       & 16 bits        & 32 bits        & 48 bits        & 64 bits        &  $\Delta$         \\
\midrule
\multirow{5}{*}{AlexNet~\cite{alexnet}}     & CNNH$^\star$~\cite{CNNH}               & 0.281         & 0.450          & 0.525          & 0.554          &          & 0.564          & 0.574          & 0.571          & 0.567          &           \\
          & DNNH$^\star$~\cite{DNNH}               & 0.290          & 0.461          & 0.530          & 0.565          &          & 0.593          & 0.603          & 0.605          & 0.610          &           \\
          & HashNet$^\star$~\cite{HashNet}            & 0.506          & 0.631          & 0.663          & 0.684          &          & 0.687          & 0.718          & 0.730          & 0.736          &  \\
          & DCCH$^\star$~\cite{DCCH}               & -              & -              & -              & -              &          & 0.659          & 0.729          & 0.731          & 0.739          &           \\
          & OrthoCos$^\star$~\cite{OrthoHash}           & 0.606          & 0.679          & -              & 0.711          &          & 0.709          & 0.762          & -              & 0.787          &           \\ \midrule
\multirow{10}{*}{GoogLeNet~\cite{googlenet}} & DPSH$^\star$~\cite{DPSH}               & -              & -              & -              & -              &          & 0.349          & 0.355          & 0.360          & 0.367          &           \\
          & DSDH$^\star$~\cite{DSDH}               & -              & -              & -              & -              &          & 0.347          & 0.359          & 0.366          & 0.370          &           \\
          & DCWH$^\star$~\cite{DCWH}               & 0.782          & 0.799          & 0.835          & 0.849          &          & 0.742          & 0.776          & 0.786          & 0.779          &           \\
          & IDCWH$^\star$~\cite{IDCH}              & -              & -              & -              & -              &          & 0.732          & 0.760          & 0.764          & 0.770          &           \\ \cmidrule{2-12} 
          & DHN$^\dagger$~\cite{DHN}                & 0.542          & 0.582          & 0.541          & 0.627          & -        & 0.677          & 0.701          & 0.695          & 0.694          & -         \\
          & DHN$^\dagger$ + HHF          & 0.721          & 0.810          & 0.812          & 0.802          & +21.33\% & 0.722          & 0.752          & 0.763          & 0.772          & +6.05\%   \\ \cmidrule{2-12} 
          & Proxy-NCA$^\dagger$~\cite{Proxy-NCA}          & 0.613          & 0.690           & 0.749          & 0.755          & -        & 0.562          & 0.608          & 0.674          & 0.649          & -         \\
          & Proxy-NCA$^\dagger$ + HHF    & 0.803          & 0.854          & 0.873          & 0.872          & +14.88\% & 0.708          & 0.839          & \underline{0.860}          & \underline{\textbf{0.870}} & +19.6\%   \\ \cmidrule{2-12} 
          & Proxy-Anchor$^\dagger$~\cite{Proxy_anchor_loss}       & 0.856          & 0.883          & 0.880           & 0.883          & -        & 0.736          & 0.822          & 0.845          & 0.850          & -         \\
          & Proxy-Anchor$^\dagger$ + HHF & \underline{\textbf{0.882}} & \underline{\textbf{0.898}} & \underline{\textbf{0.901}} & \underline{\textbf{0.897}} & +1.90\%  & \underline{\textbf{0.805}} & \underline{\textbf{0.850}} & 0.858          & 0.851          & +2.78\%   \\ \midrule
\multirow{9}{*}{ResNet-50~\cite{ResNet}}  & DCH$^\star$~\cite{DCH}                & 0.652          & 0.737          & -              & 0.758          &          & 0.759          & 0.801          & -              & 0.825          &           \\
          & CSQ$^\star$~\cite{CSQ}                & 0.851          & 0.865          & 0.866              & 0.873          &          & 0.796          & 0.838          & -              & 0.861          &           \\
          & DCSH$^\star$~\cite{DCSH}               & -              & -              & -              & -              &          & \underline{\textbf{0.805}} & 0.847          & 0.859          & 0.861          &           \\ \cmidrule{2-12} 
          & DHN$^\dagger$~\cite{DHN}                & 0.429          & 0.501          & 0.516          & 0.552          & -        & 0.650          & 0.663          & 0.667          & 0.667          & -         \\
          & DHN$^\dagger$ + HHF          & 0.653          & 0.769          & 0.799          & 0.807          & +25.75\% & 0.746          & 0.798          & 0.814          & 0.826          & +13.43\%  \\ \cmidrule{2-12} 
          & Proxy-NCA$^\dagger$~\cite{Proxy-NCA}          & 0.620          & 0.715          & 0.737          & 0.769          & -        & 0.545          & 0.609          & 0.659          & 0.679          & -         \\
          & Proxy-NCA$^\dagger$ + HHF    & 0.835          & 0.861          & 0.859          & 0.858          & +14.3\% & 0.700          & 0.796          & 0.850          & 0.857          & +17.78\%  \\ \cmidrule{2-12} 
          & Proxy-Anchor$^\dagger$~\cite{Proxy_anchor_loss}       & 0.824          & 0.871          & 0.874          & 0.873          & -        & 0.713          & 0.805          & 0.840          & 0.846          & -         \\
          & Proxy-Anchor$^\dagger$ + HHF & \underline{0.871}          & \underline{0.891}          & \underline{0.895}          & \underline{0.896}          & +2.78\%  & 0.798          & \underline{\textbf{0.850}} & \underline{\textbf{0.872}} & \underline{\textbf{0.870}} & +4.65\%    \\
\bottomrule
\end{tabular}}
\vspace{-16pt}
\label{Tab.ImageNet_COCO}
\end{table*}

\textbf{Settings.}
We integrate Proxy-Anchor Loss~\cite{Proxy_anchor_loss} with HHF and quantization terms for experiments. The overall loss function has illustrated in Eq.~\ref{Eq.total_loss}. To eliminate the influence of backbones on the performance, we fairly compare various methods on CIFAR~\cite{cifar} datasets with the same GoogLeNet for all methods. In Tab.~\ref{Tab.CIFAR}, ``$^\star$" represents that we directly report the results from~\cite{DCWH,IDCH}. DPSH~\cite{DPSH}, DTSH~\cite{DTSH}, and DSDH~\cite{DSDH} are reproduced by IDCH~\cite{IDCH} on GoogLeNet, and we directly adopt the results from~\cite{IDCH}. We also reproduce and report the results of Proxy-Anchor Loss with quantization term on GoogLeNet for reference (denoted as ``$^\dagger$").

\textbf{Results.}
In simple CIFAR-10, the conflict gets relieved by expanding train data under little category numbers. The bottleneck originates from the expressiveness of backbones, which accounts for the similar performance on CIFAR-10-FULL between Proxy-Anchor with/without HHF. Tab.~\ref{Tab.CIFAR} demonstrates that HHF outperforms existing techniques over different hash bits, especially in CIFAR-100-FULL. Since CIFAR-100 has more categories than CIFAR-10, the conflict between metric learning and quantization learning is more prominent and challenging. In CIFAR-100-FULL, HHF achieves over $\bm{5.78\%}$ and $\bm{2.35\%}$ average gains compared to IDCWH~\cite{IDCH} and Proxy-Anchor~\cite{Proxy_anchor_loss}, respectively, which verifies the effectiveness of the proposed method.

\begin{table*}[ht!]
\setlength{\tabcolsep}{6pt}
\centering
\caption{Ablation study with proxy-based methods on ImageNet under ResNet-50. Underlined results: Smaller HPE, $\eta_{\text{global}}$, and $\eta_{\text{local}}$, which indicate better performance. HHF enables to achieve better HPE without sacrificing metric learning. }
\vspace{-6pt}
\normalsize
\resizebox{1\linewidth}{!}{%
\begin{tabular}{l|lll|lll|lll|lll}
\toprule
Hash Bits           & \multicolumn{3}{c|}{16 bits} & \multicolumn{3}{c|}{32 bits} & \multicolumn{3}{c|}{48 bits} & \multicolumn{3}{c}{64 bits} \\ \midrule
Metric  & HPE $\downarrow$    & $\eta_{\text{global}} \downarrow$   & $\eta_{\text{local}} \downarrow$     & HPE $\downarrow$    & $\eta_{\text{global}} \downarrow$   & $\eta_{\text{local}} \downarrow$    & HPE$ \downarrow$    & $\eta_{\text{global}} \downarrow$   & $\eta_{\text{local}} \downarrow$  & HPE$ \downarrow$   & $\eta_{\text{global}} \downarrow$   & $\eta_{\text{local}} \downarrow$   \\ \midrule
Proxy-NCA          & 1.858 & 0.298 & 0.947 & 2.519 & 0.286 & 0.806 & 2.875 & 0.281 & 0.748 & 3.039 & 0.263 & 0.679 \\
Proxy-NCA + HHF    & \underline{1.334} & \underline{0.262} & \underline{0.466} & \underline{1.690} & \underline{0.250} & \underline{0.444} & \underline{2.052} & \underline{0.243} & \underline{0.438} & \underline{2.250} & \underline{0.235} & \underline{0.450} \\ \midrule
Proxy-Anchor       & 1.862 & 0.295 & 0.434  & 1.834 & 0.291  & 0.413 & 2.072 & 0.285  & 0.406 & 2.300 & 0.282 & 0.406  \\
Proxy-Anchor + HHF & \underline{1.296} & \underline{0.265} & \underline{0.433} & \underline{1.807} & \underline{0.267}  & \underline{0.407} & \underline{1.846} & \underline{0.264} & \underline{0.399} & \underline{2.103} & \underline{0.262}	& \underline{0.393}   \\ 
\bottomrule
\end{tabular}}
\label{Tab.Further_Analyses}
\end{table*}

\begin{figure*}[ht!]
  \vspace{-10pt}
  \centering
  \includegraphics[width=1.0\linewidth]{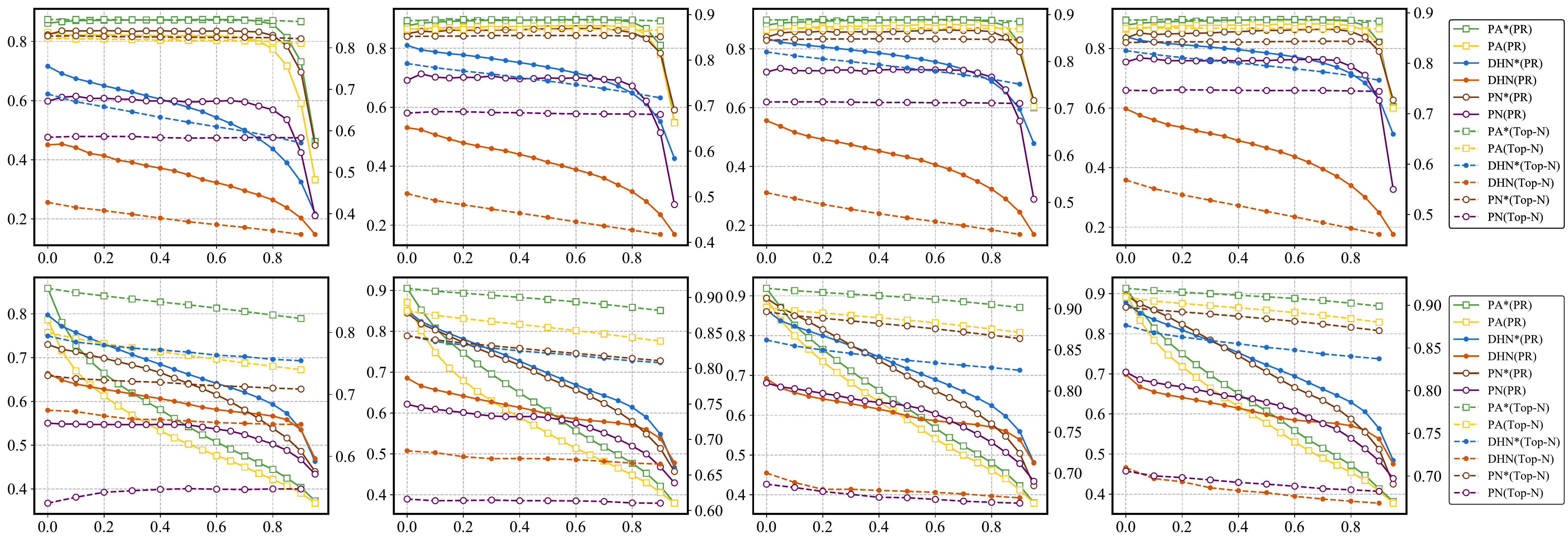}
  \vspace{-20pt}
  	\caption{Precision-recall curves (solid lines) and Precision@top-$N$ curves (dot lines) on ImageNet (@first row) and MS-COCO (@second row) under ResNet-50 with binary codes @16 / 32 / 48 / 64 bits (left $\rightarrow$ right). ``*" indicates method with HHF. PA: Proxy-Anchor. PN: Proxy-NCA. Three baselines can be further improved with our HHF.}
  	\label{Fig.Recall}
  	\vspace{-20pt}
\end{figure*}

\vspace{-8pt}
\subsection{Results on Large-Scale Datasets}

\textbf{Settings.}
To illustrate the effectiveness and generalization of HHF comprehensively, we integrate HHF with three different methods using different backbones on challenging large-scale datasets (i.e., ImageNet and MS-COCO). Specifically, we adopt HHF in 1). DHN~\cite{DHN}, a pair-based loss, combines quantization loss with metric loss to constrain latent codes. 2). Proxy-NCA~\cite{Proxy-NCA}, a proxy-based loss, proposes a set of learnable class centers. 3). Proxy-Anchor~\cite{Proxy_anchor_loss}, a proxy-based loss, combines the advantages of proxy-based methods and pair-based methods. We comprehensively compare them to various state-of-the-art methods in Tab.~\ref{Tab.ImageNet_COCO}.

\textbf{Results.} Tab.~\ref{Tab.ImageNet_COCO} demonstrates that HHF enables different methods to achieve significant performance gains in the two large-scale datasets over different hash bits. As mentioned before, the conflict issue becomes increasingly severe as the category increases, HHF keeps generalizations on challenging datasets and shows portability in many approaches of dealing with conflicts, especially in the na\"ive methods DHN and Proxy-NCA. Experiments on different backbones also verify that HHF is robust and effective in tackling the conflict between metric learning and quantization learning.

\vspace{-10pt}
\subsection{Ablation study.}

\textbf{Qualitative analysis.}
Fig.~\ref{Fig.Recall} shows the Precision w.r.t. different numbers of Top Returned Samples and Recall curves on ImageNet and MS-COCO. Compared to methods with/without HHF, HHF achieves considerable performance improvements compared with original ones. Specifically, Proxy-Anchor with HHF achieves the best performance among others.

\textbf{Quantitative analysis.}
As one of the main methodologies of HHF, we propose the hinge inflection point to explicitly constrain the metric loss to avoid class centers becoming unlimited alienated from each other, hence HHF seems to damage the results of feature learning. However, thanks to the theory of minimum distance of the binary linear code~\cite{codetable}, we quantitatively demonstrate that HHF generally will not damage metric learning, instead it slightly promotes the metric learning process and ensures clusters more distinguishable, because it guarantees metric loss when distance $\le \zeta$ and fully focuses on quantization learning if distance $\ge \zeta$.

Tab.~\ref{Tab.Further_Analyses} shows that the state-of-the-art proxy-based methods keep stable $\eta$ with/without HHF, i.e., HHF has seldom side-effect on metric learning (as we will verify in Fig.~\ref{Fig.tsne}). Noteworthy, methods with HHF get HPE significantly decreased, i.e., the latent codes are closer to the ideal binary position, which is crucial for efficient image retrieval. In most cases, HHF enables distinguishing confusion categories, i.e., it gets $\eta_{\text{local}}$ significantly decreased. The result proves that the proposed HHF generally reduces semantic information loss in the binarization process without loss of cluster performance.

\begin{figure*}[ht!]  
\centering
    \subfloat[Query: doormat]{
        \begin{minipage}[b]{0.48\textwidth}
            \includegraphics[width=1\textwidth]{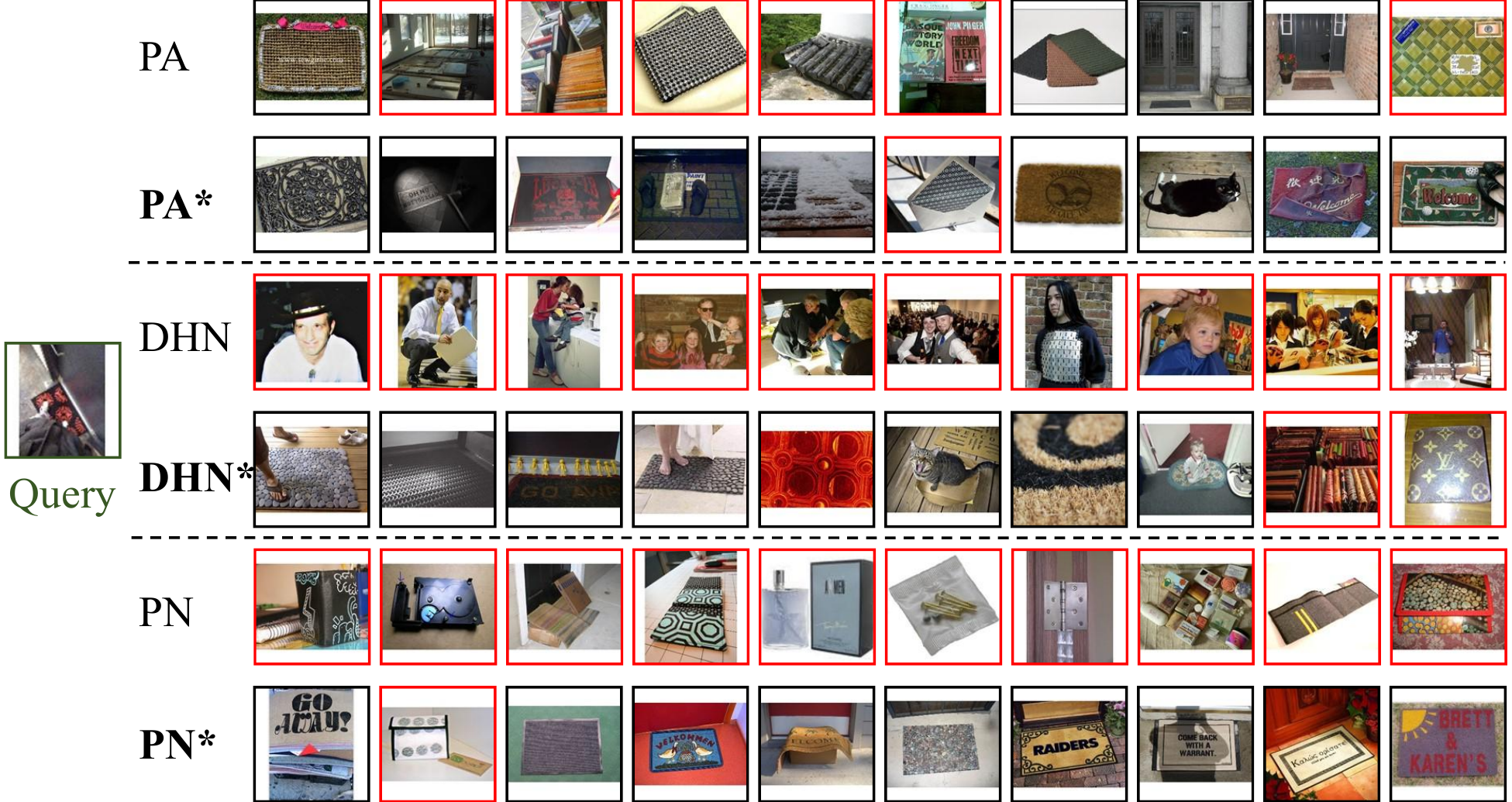}
        \end{minipage}
    }
    \hfill	
    \subfloat[Query: butterfly]{
        \begin{minipage}[b]{0.48\textwidth}
            \includegraphics[width=1\textwidth]{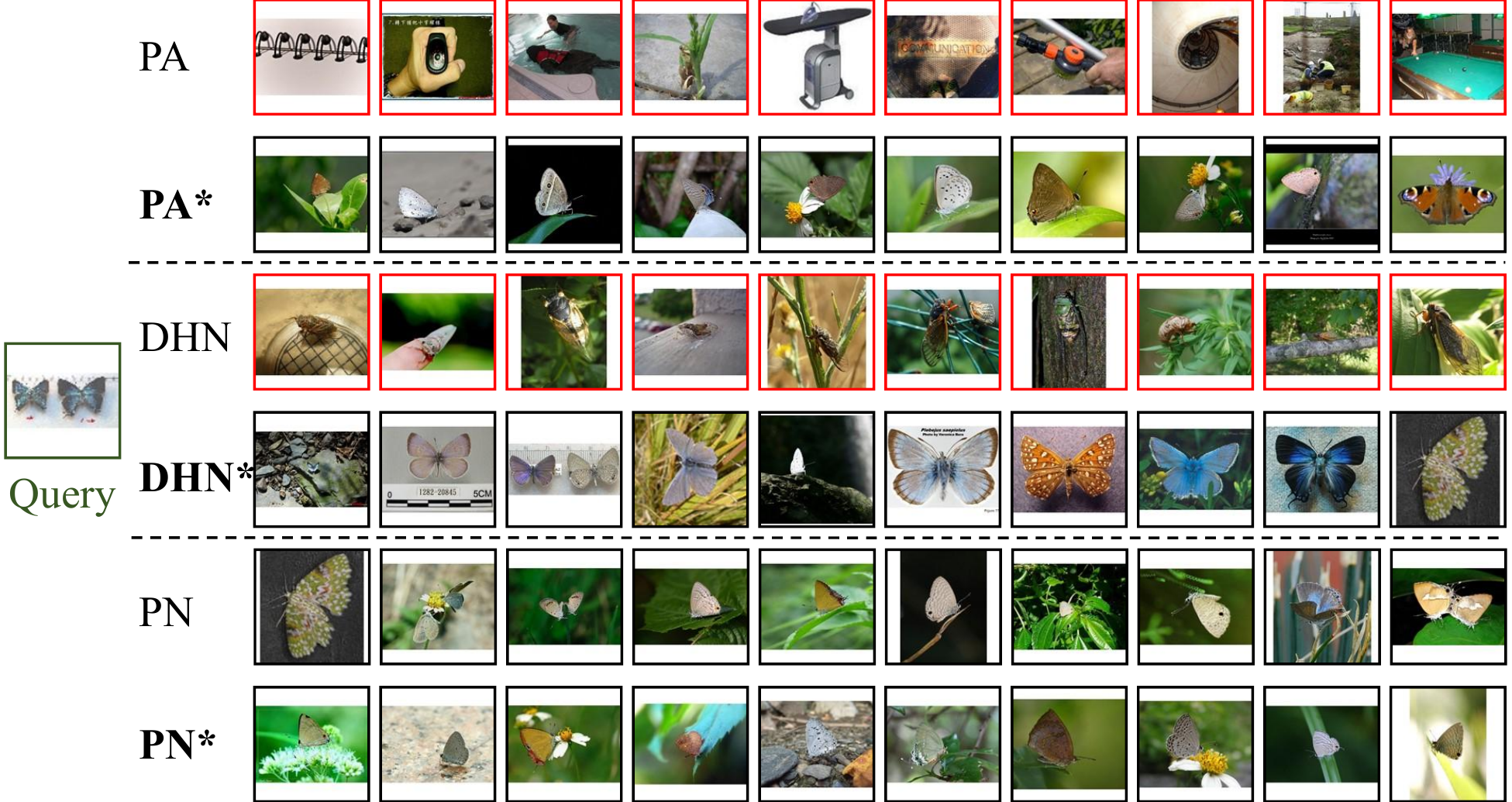}
        \end{minipage}
    }
    \vspace{-6pt}
    \caption{Retrieval results of methods with/without HHF on ImageNet. Red box indicates misclassified results, while HHF enables to significantly improve the retrieval quality. ``*" indicates method with HHF. PA: Proxy-Anchor. PN: Proxy-NCA.}
    \label{Fig:retrieval1}
    \vspace{-15pt}
\end{figure*}

\vspace{-8pt}
\subsection{Visualized Comparison}
\textbf{Retrieval results.}
We represent visualized retrieval results for empirical comparisons. We randomly give some queries to retrieve the top-$10$ results from ImageNet database and compare the retrieved images of different methods (i.e., DHN, Proxy-NCA, Proxy-Anchor with/without HHF). The results are illustrated in Fig.~\ref{Fig:retrieval1}.

Results in Fig.~\ref{Fig:retrieval1} show that the proposed HHF enables to improve the retrieval performance significantly. Specifically, due to the conflict between metric learning and quantization learning, previous methods cannot balance the optimal clustering results and ideal hashing position. Intuitively, although some methods with/without HHF perform comparable on clustering (i.e., similar $\eta$) in Tab.~\ref{Tab.Further_Analyses}, there are many misclassified or error hashed results in retrieval when without HHF (e.g., for a butterfly query, Proxy-Anchor misclassify the results and make all error retrieval results; for a doormat query, DHN retrieve the all error human results).

In contrast, with the combination of the proposed HHF, the retrieval accuracy gets remarkably improved. HHF enables images of the same category to accurately binarize into similar hash codes (i.e., smaller HPE and larger mAP), which alleviates the errors with satisfactory retrieval results and is robust in most cases. Intuitively, the retrieval results in Fig.~\ref{Fig:retrieval1} are more reliable than those without HHF.

\begin{figure}[t!]
    \centering
    \vspace{-10pt}
    \subfloat[DHN~\cite{DHN}]{
    
        \includegraphics[width=0.31\linewidth]{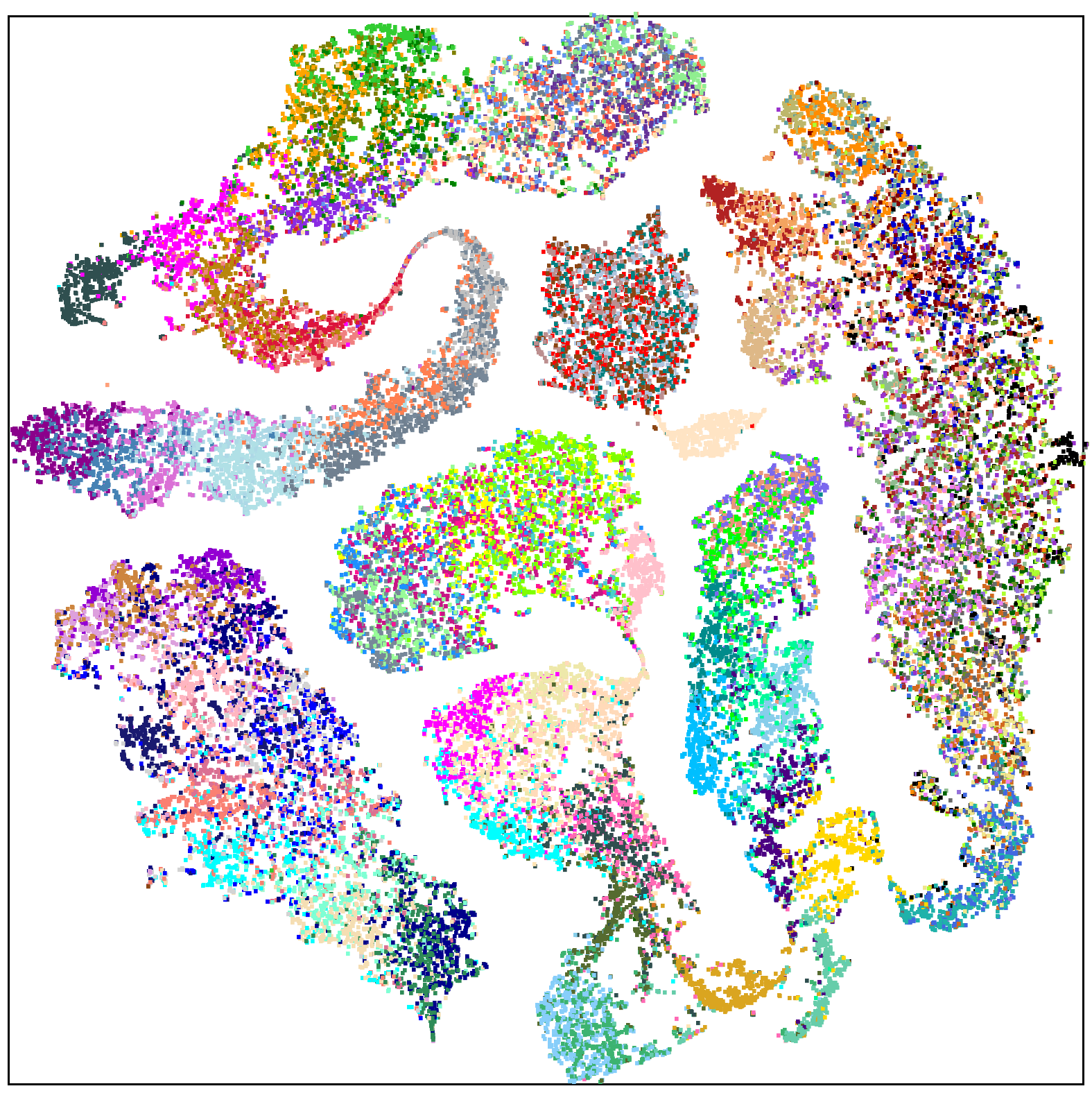}
    }
    \subfloat[Proxy-NCA~\cite{Proxy-NCA}]{
        \includegraphics[width=0.31\linewidth]{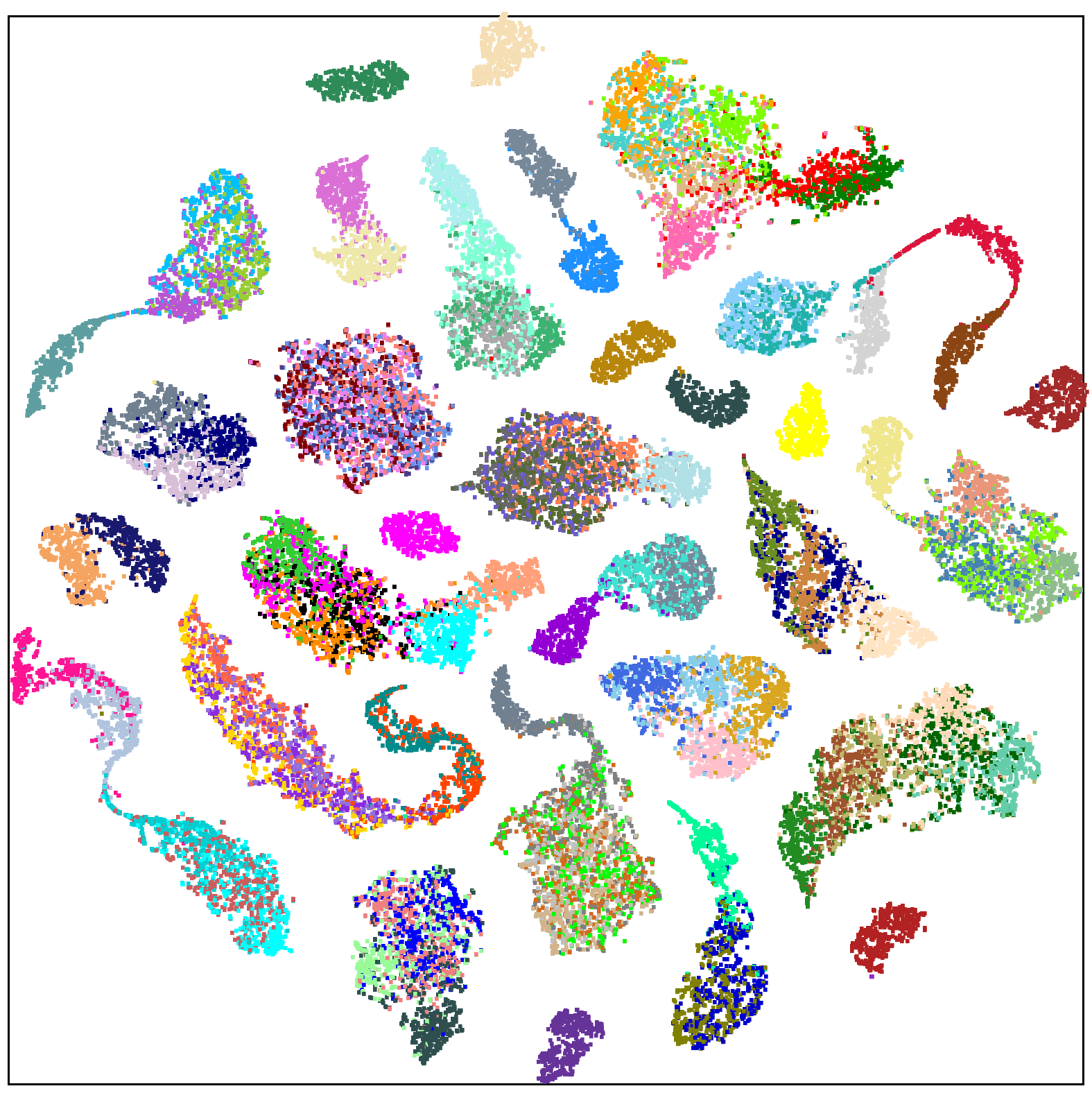}
    }
    \subfloat[Proxy-Anchor~\cite{Proxy_anchor_loss}]{
        \includegraphics[width=0.31\linewidth]{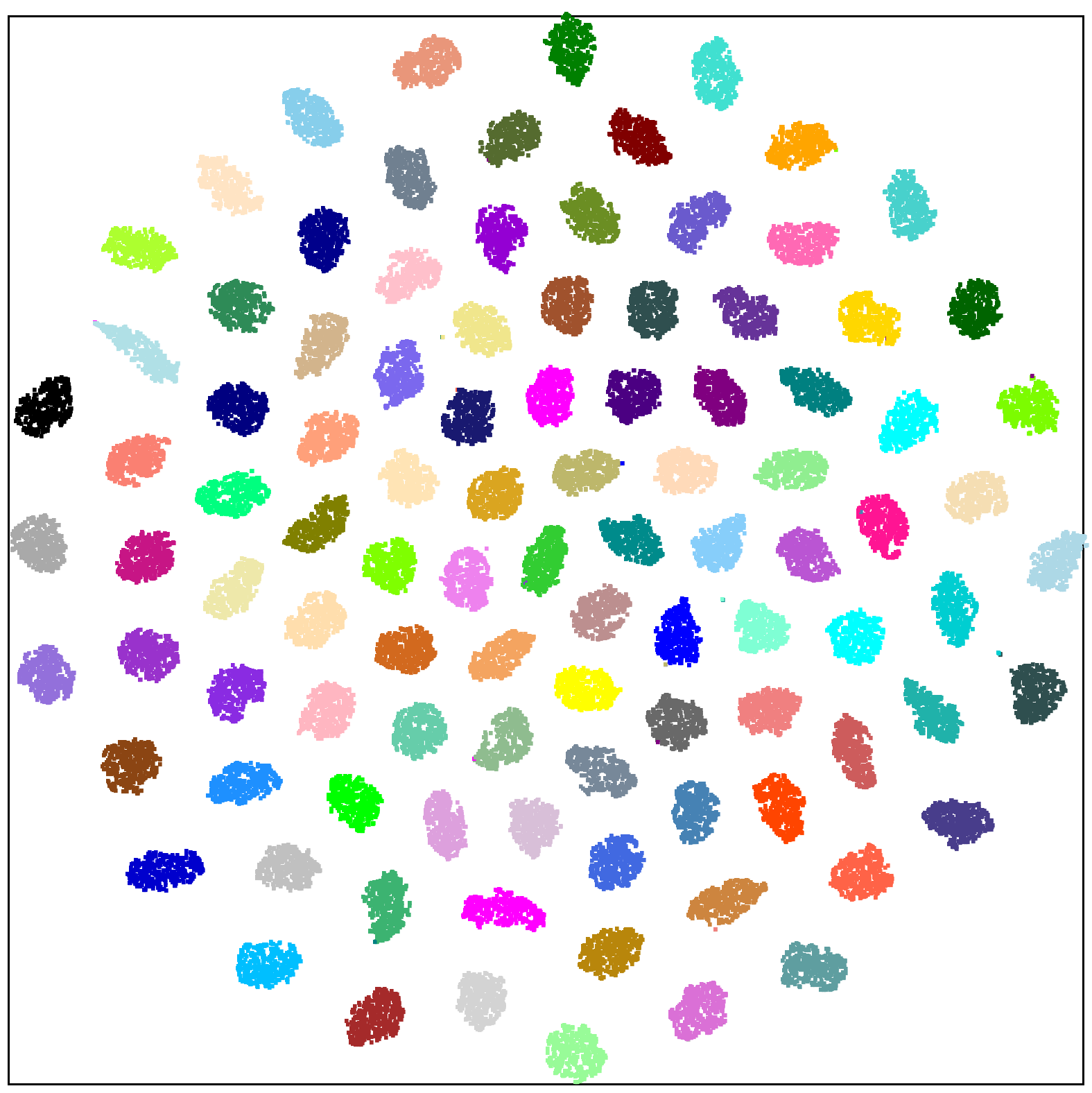}
    }
    \vspace{-12pt}
    \hfill
    \subfloat[\textbf{DHN+HHF}]{
        \includegraphics[width=0.31\linewidth]{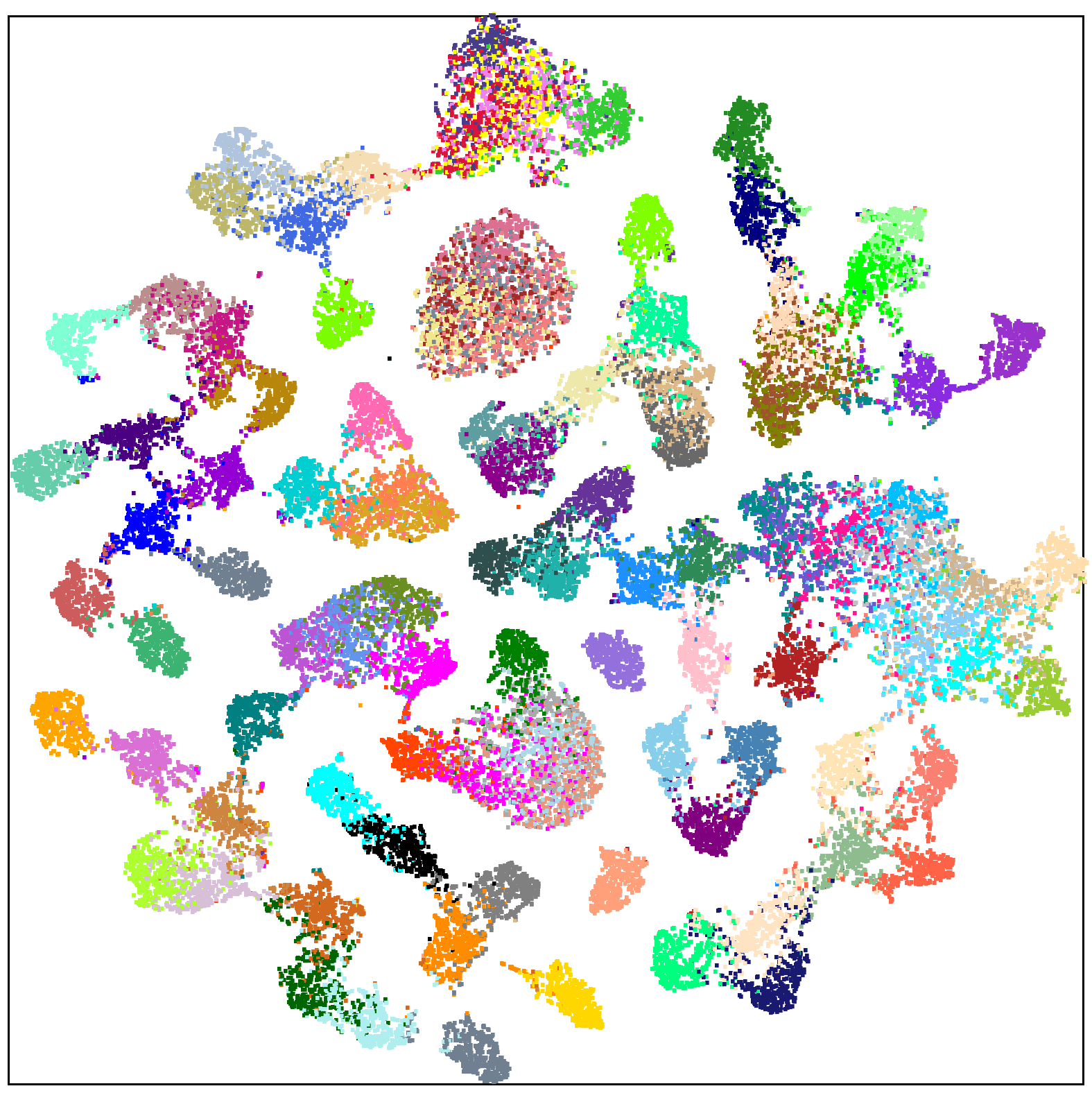}
    }
    \subfloat[\textbf{Proxy-NCA+HHF}]{
        \includegraphics[width=0.31\linewidth]{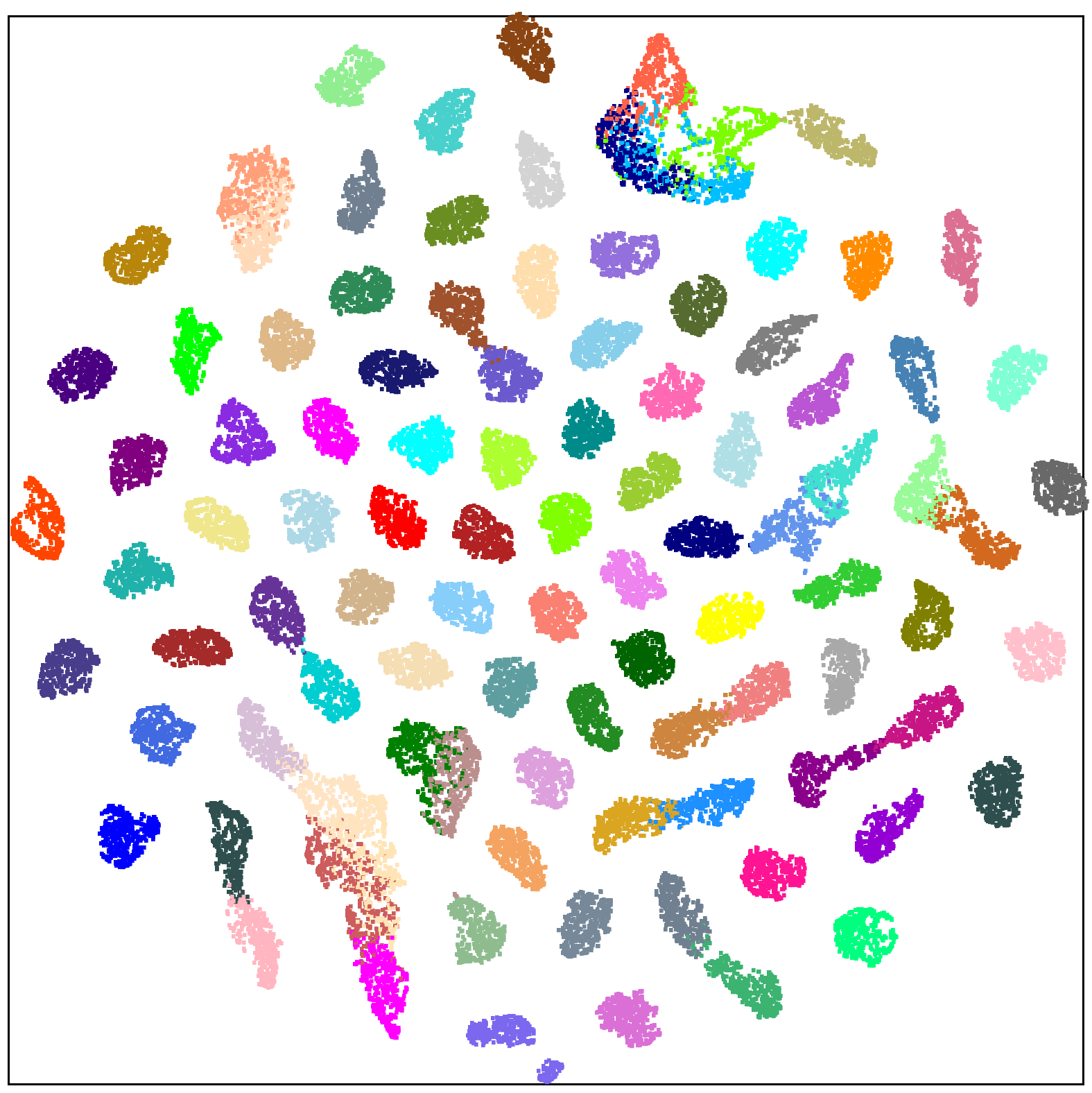}
    }
    \subfloat[\textbf{Proxy-Anchor+HHF}]{
        \includegraphics[width=0.31\linewidth]{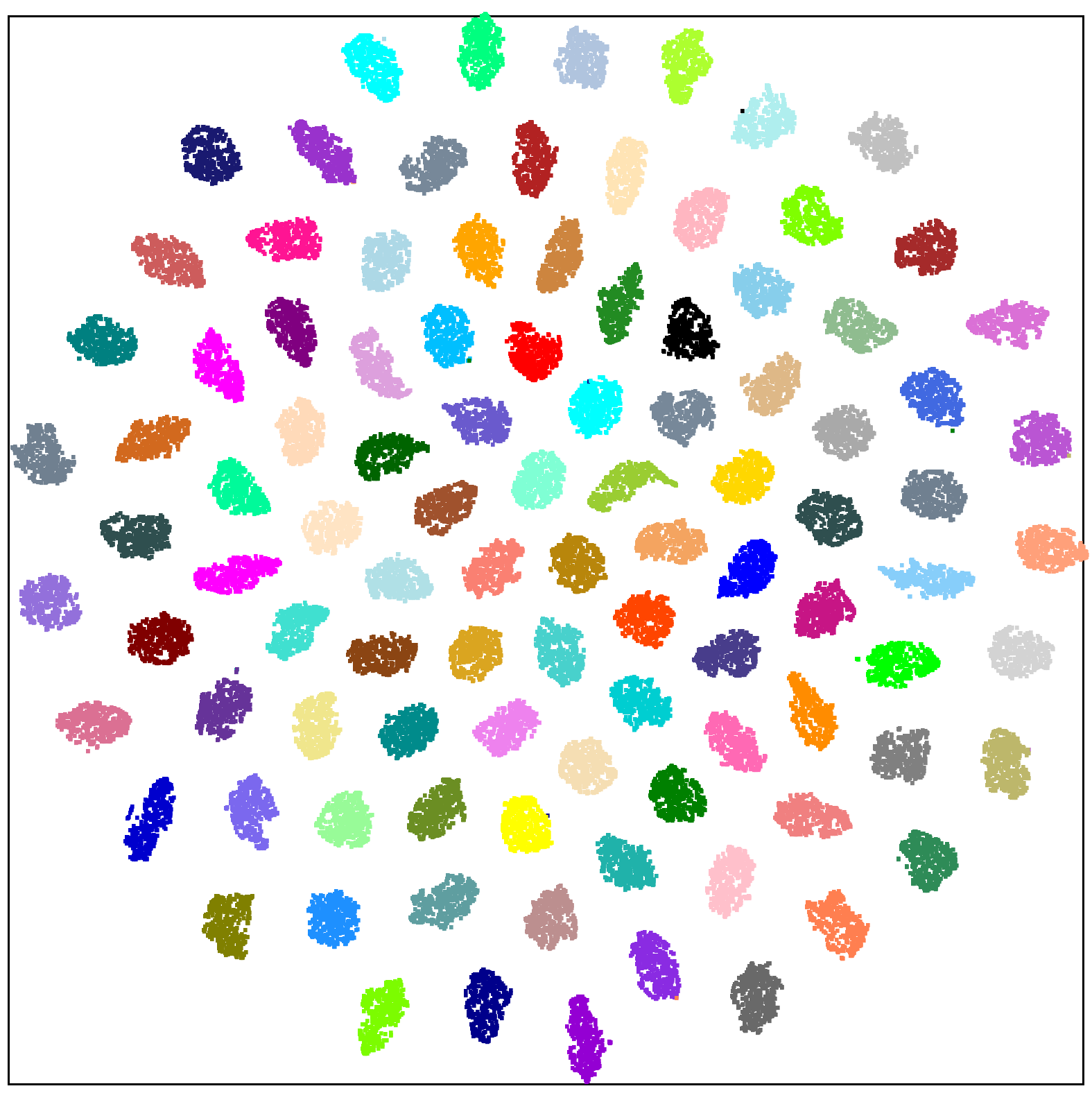}
    }%
    \centering
    \vspace{-6pt}
    \caption{Visualized t-SNE results of methods with/without HHF on CIFAR-100. The scatters of the same color indicate the same categories, the combination of HHF enables to disentangle different categories.}
    \label{Fig.tsne}
    \vspace{-20pt}
\end{figure}

\textbf{Clustering results.}
To fully illustrate how the proposed HHF contributes to metric learning, besides the detailed quantitative comparisons in Tab.~\ref{Tab.Further_Analyses}, we further present our empirical study to demonstrate that HHF will not harm metric learning. Specifically, we utilize t-SNE~\cite{TSNE} to map $K$-dimension latent codes into 2D distribution for visualization. We compared DHN~\cite{DHN}, Proxy-NCA~\cite{Proxy-NCA}, and Proxy-Anchor~\cite{Proxy_anchor_loss} with/without HHF on CIFAR-100 with $12$ hash bits, as illustrated in Fig.~\ref{Fig.tsne}.

Fig.~\ref{Fig.tsne}(a) shows that the latent codes of different categories are coupled together, indicating that DHN is incapable of achieving satisfactory clustering results. The poor clustering performance also deteriorates the subsequent quantization operation, making the retrieval results less promising (as Fig.~\ref{Fig:retrieval1} illustrates). Noteworthy, DHN uses a pair-wise loss for metric learning. However, such pair-wise loss methods only consider a negative sample once in back-propagate, regardless of other negative ones. Hence, the randomly sampled pairs cannot guarantee that each category is alienated and harmful to other negative samples, making it easy to trap in a locally optimal solution. In contrast, HHF makes specific constraints to avoid samples being over-alienated, which tackles the conflict of intra-class and inter-class.

Compared to DHN, Proxy-NCA slightly disentangles the hash centers of different categories. However, most categories are still ambiguous of each other. As a comparison, Proxy-NCA with HHF significantly improves both metric learning (Fig.~\ref{Fig.tsne}(e)) and quantization learning (Tab.~\ref{Fig:retrieval1}). Proxy-Anchor is proved to be helpful in metric learning. Intuitively, the latent codes of each category can get perfectly disentangled compared to DHN. Fig.~\ref{Fig:retrieval1}(c)(f) demonstrate that both Proxy-Anchor with and without HHF achieve appealing results. HHF successfully solves the challenging conflict and improves retrieval performance without losing of metric learning. 

\begin{table}[t!]
	\caption{mAP performance of HHF@48bits w.r.t. the margin hyperparameter $\delta$ on three datasets under GoogLeNet, Above which the performance is optimal when $\delta = 0.2$.}
	\vspace{-6pt}
\setlength{\tabcolsep}{7pt}
\centering
\resizebox{1\linewidth}{!}{%
\begin{tabular}{l|ccccc}
\toprule
Margin Params $\delta$ & 0.0   & 0.1   & 0.2   & 0.3  &  0.4 \\ \midrule
CIFAR-100-FULL     & 0.818 & 0.858 & \textbf{0.875} & 0.874 & 0.367\\
ImageNet     & 0.879 & 0.889 & \textbf{0.901} & 0.879 & 0.420 \\
MS-COCO         & 0.841 & 0.847 & \textbf{0.858} & 0.837 & 0.541 \\ 
\bottomrule
\end{tabular}}
  	\label{Tab.delta}
  	\vspace{-20pt}
\end{table}

\vspace{-15pt}
\subsection{Hyperparameter Sensitivity Analysis}

\textbf{Settings.}
We further present in-depth investigations on the influence of $\delta$. The main effect of $\delta$ in the image retrieval field is to prevent overfitting~\cite{Triplet}. $\delta$ shows more interpretability in HHF that it prevents different images of the same category encoded into the same latent codes. It ensures images to distribute in an ideal hypersphere. We comprehensively compare Proxy-Anchor with HHF setting the margin $\delta$ range from $0$ to $0.4$. We conduct experiments in three standard datasets: CIFAR-100, ImageNet, and MS-COCO with hash bits of $48$. The mAP results are illustrated in Tab.~\ref{Tab.delta}.

\textbf{Results.}
Tab.~\ref{Tab.delta} illustrates that $\delta$ has similar performance influence trends on these benchmarks of a specific approach, where the best performance achieves nearby $\delta = 0.2$. We also report HHF without adding $\delta$ for reference, to intuitively demonstrate that the missing margin will generally damage the performance. 
Specifically, due to the existence of overfitting issues, mAP of each dataset gets decreased compared to the best results. Noteworthy, the improving $\delta$ does not guarantee the consistent performance gains of different methods. When $\delta$ is away from $0.2$, we notice that performance will oppositely decrease. Hence the empirical study also proves the importance of an appropriate $\delta$ for better hash performance.

\section{Conclusion} \label{sec:conclusion}

In this paper, a novel hinge loss function is proposed towards the ideal hash position of given images for large-scale retrieval, which is robust to plugged-and-play into existing methods to consistently improve their performance. 
To achieve this, we explore in-depth investigations to excavate the relationship between metric learning and quantization learning, and systematically analyze the challenging conflict issues between the two learning procedures for deep hashing. By revealing the unsolved fact that the ideal metric solution cannot satisfy the optimal quantization solution for retrieval, we focus on addressing semantic information loss during the hashing process in large-scale image datasets and carefully design Hashing guided hinge loss to effectively tackle such conflict.
Extensive experiments on four standard datasets verify the superiority and flexibility of HHF to integrate with various off-the-shelf methods to achieve significant performance gains. We conduct both quantitative and qualitative comparisons to demonstrate the effectiveness of our method. Various evaluation metrics and visualizations fully demonstrate that HHF enables to achieve
state-of-the-art retrieval performance with better visually consistent results.

\bibliographystyle{IEEEtran}
\bibliography{main}

\end{document}